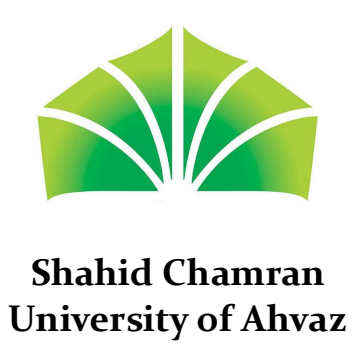

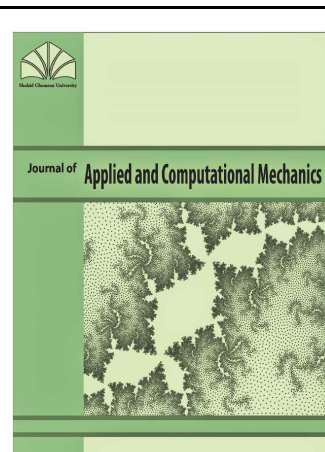

# Time Distance: A Novel Collision Prediction and Path Planning Method

Ali Analooee[1], Shahram Azadi[2], Reza Kazemi[3]

[1] Department of Mechanical Engineering, K. N. Toosi University of Technology, Tehran, 1991943344, Iran, Email: aanalooee@mail.kntu.ac.ir

[2] Department of Mechanical Engineering, K. N. Toosi University of Technology, Tehran, 1991943344, Iran, Email: azadi@kntu.ac.ir

[3] Department of Mechanical Engineering, K. N. Toosi University of Technology, Tehran, 1991943344, Iran, Email: kazemi@kntu.ac.ir

Corresponding author: S. Azadi (azadi@kntu.ac.ir)

**Abstract.** In this paper, a new fast algorithm for path planning and a collision prediction framework for two dimensional dynamically changing environments are introduced. The method is called Time Distance (TD) and benefits from the space-time space idea. First, the TD concept is defined as the time interval that must be spent in order for an object to reach another object or a location. Next, TD functions are derived as a function of location, velocity and geometry of objects. To construct the configuration-time space, TD functions in conjunction with another function named "Z-Infinity" are exploited. Finally, an explicit formula for creating the length optimal collision free path is presented. Length optimization in this formula is achieved using a function named "Route Function" which minimizes a cost function. Performance of the path planning algorithm is evaluated in simulations. Comparisons indicate that the algorithm is fast enough and capable to generate length optimal paths as the most effective methods do. Finally, as another usage of the TD functions, a collision prediction framework is presented. This framework consists of an explicit function which is a function of TD functions and calculates the TD of the vehicle with respect to all objects of the environment.

## 1. Introduction

Path planning is an innovation-demanding field in control engineering and robotics. Research in this area has yielded plenty of remarkable methods. These methods are divided into classical and heuristic algorithms. Classical algorithms include cell decomposition, potential field, roadmap and sampling-based methods.

In cell decomposition methods, the path is obtained by connecting a set of simple cells which decompose the free space of the robot's configuration space. The research presented by Nilsson [1] is the leading originator of the cell decomposition idea. Detailed explanations of these methods are represented in [2-4]. As a powerful method widely used in robotics, artificial potential field (APF) [5-9] has been developed by many researchers thus far. This method has also been utilized for path planning of automated ground vehicles. As they are non-scenario based, APF methods are advantageous in comparison to most methods associated with path planning for ground vehicles. The works presented by Kala and Warwik [10], Rasekhipour et al. [11, 12] and Ji et al. [13] are some instances of this type. In sampling-based motion planning methods, the configuration space is sampled by some random or quasi-random points, and a search algorithm is used to find a path by connecting the sample points. In spite of being fast for complex problems, these algorithms are suboptimal. The most popular sampling-based methods are probabilistic roadmap (PRM) [14-19] and rapidly exploring random trees (RRT) [20]. Comprehensive studies of sampling-based methods are available in [21] and [22]. In roadmap methods, the free configuration space is mapped into a network of one-dimensional lines, and a search algorithm is applied to this network to find the shortest path. The two most important roadmap methods include visibility graph [1, 23, 24] and Voronoi diagram [25]. To study these methods, we recommend the readers to see [2, 4].

Heuristic algorithms are increasingly becoming popular and desirable. Most of these algorithms are inspired by biological behaviors. The three well-known, nature-inspired algorithms are genetic algorithm (GA) [26], particle swarm optimization (PSO) [27], and ant colony optimization (ACO) [28-30]. A comprehensive review of neural network, fuzzy logic, nature-inspired methods and hybrid algorithms is presented by Mac et al. [31].

### 1.1 The space-time space

The idea of path planning in the space-time space was presented by Erdmann and Lozano-Perez [32] by introducing the configuration-time space. This concept greatly facilitates collision detection in dynamically changing environments.

Cefalo and Oriolo [33] used the configuration-time space and state-time space for collision checking in a motion planning framework. In this study, collision checking was performed using a randomized search algorithm. Schwesinger et al. [34] used workspace-time space to check the candidate trajectories of a sampling-based motion planner for collision with dynamic obstacles. In this study, which benefits from bounding volume hierarchy data structures, axis-aligned bounding boxes were used to detect collisions. Xidias and Aspragathos [35] also exploited the workspace-time concept to present a three phase path planning framework. In this approach, the workspace-time space was constructed in the first phase. Then, the Bump-Surface concept and the genetic algorithm were used to generate length optimal trajectories for a set of mobile robots in a dynamic environment. Alonso-Mora et al. [36] used workspace-time idea to present a distributed formation planning method for a team of ground or aerial robots. The method was capable of reconfiguring the formation of the robots to avoid collision with static and moving obstacles. Ragaglia et al. [37] exploited this idea to present a novel trajectory generation algorithm to take into account the safety requirements for an industrial robot involved in a human-robot collaboration. They calculated swept volumes to model the space occupied by the human worker's body in a limited time predicted by a sensor fusion strategy. Then, the trajectory generation algorithm modified the pre-programmed trajectory to prevent collisions between the robot and the worker considering the swept volumes as obstacles. Gaschler et al. [38] presented a task and motion planning framework named KABouM. They introduced a collision detection algorithm by defining bounded geometric predicates for collision and inclusion. They also proposed a set of algorithms called bounding mesh algorithm, bounded convex decomposition, and swept volume generation which were compatible with the definition of predicates and were used to evaluate them. In this study, to generate swept volumes, first, the path is discretized. Then, the swept volume is generated from subsequent convex hulls which are computed by applying a forward kinematics function.

### 1.2 Contributions

In this paper, a novel method for path planning in two dimensional dynamic environments is developed. The most important feature of this method is its quickness which stems from the fact that this framework uses an explicit formula to generate the path. This algorithm was shortly introduced in our previous work [39], and is presented comprehensively in this paper.

The method is called Time Distance (TD) and its bases returns to the space-time space concept. First, we define TD as the time interval that must be spent in order for objects to reach each other or locations. Then, TD functions with TD as a function of objects' location, velocity, and geometry are derived. TD functions together with another function called Z-Infinity construct the vehicle's configuration-time space. Next, a function named Route Function is presented. This function, provides length optimality in the path by minimizing a cost function. Finally, the length optimal collision free path is obtained from an explicit function which is a function of TD, Z-Infinity and Route functions. Note that the formulation can be simplified for static environments. The performance of the framework is compared with some popular algorithms in simulations. The results show promising performance in terms of quickness and length optimization for the proposed method.

After introducing the path planning framework, a collision prediction algorithm is presented by exploiting the TD concept. In this algorithm, time to collision is obtained from an explicit function which is a function of TD functions.

## 2. The Time Distance Concept

In this section, first, the Time Distance (TD) concept is defined. Then, TD functions are derived for the one and two dimensional spaces.

Definition: As a relative concept, the TD between two objects, or the TD of two objects with respect to each other, is the time that must be spent in order for these two objects to touch each other. If these two objects never touch each other, the value of TD between them is $+\infty$ .

This is a definition that provides the basis for constructing our framework.

### 2.1 One dimensional space

To obtain a function that expresses TD for a one dimensional space, consider two mobile points A and B on an axis $y$ . Two kinematic equations Eq. (1) and Eq. (2) denote the location of these two mobile points as a function of time. By subtracting these two equations, the difference between the locations of A and B is obtained in Eq. (3). In a moment that the locations of these two points equal, the left side of Eq. (3) vanishes, and Eq. (4) is obtained. Assuming $\upsilon_{B/A}$ (the relative velocity of B with respect to A) to be constant, Eq. (5) is feasible. By solving Eq. (5) for $\Delta t$ , an approximation for the TD between A and B is obtained in Eq. (6). Equation (6) does not satisfy the definition of TD because in situations that $\upsilon_{B/A}$ is zero, $\Delta t$ is $\pm\infty$ (not exclusively $+\infty$ ), and in situations that these two points are getting far from each other, $\Delta t$ is negative. Equation (7) modifies Eq. (6) and presents a function that satisfies the definition of TD. This function is named TD function of B with respect to A.

$$y_A(t) = y_A + \int \upsilon_A dt \tag{1}$$

$$y_B(t) = y_B + \int \upsilon_B dt \tag{2}$$

$$y_B(t) - y_A(t) = y_B - y_A + \int (\upsilon_B - \upsilon_A) dt \tag{3}$$

$$y_B - y_A = -\int \upsilon_{B/A} dt \tag{4}$$

$$y_B - y_A = -\upsilon_{B/A} \Delta t \tag{5}$$

$$\Delta t = \frac{y_A - y_B}{\upsilon_{B/A}} \tag{6}$$

$$TD_{B/A} = 2\left(\mathrm{sign}\left(\frac{y_A - y_B}{\upsilon_{B/A}}\right) + 1\right)^{-1} \left|\frac{y_A - y_B}{\upsilon_{B/A}}\right| \tag{7}$$

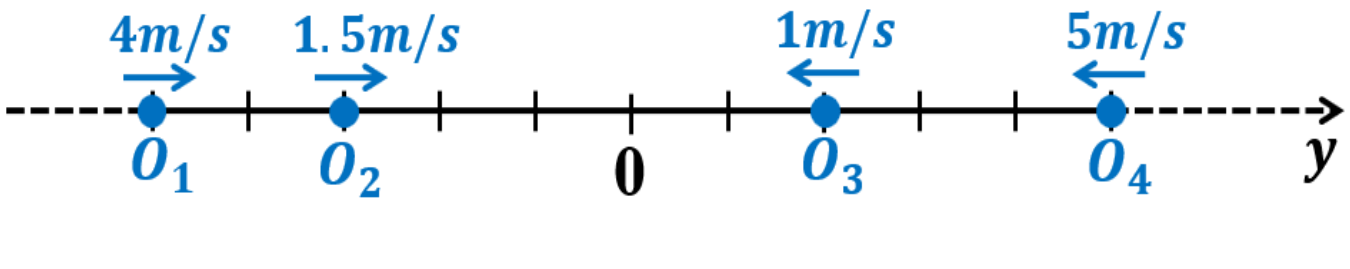

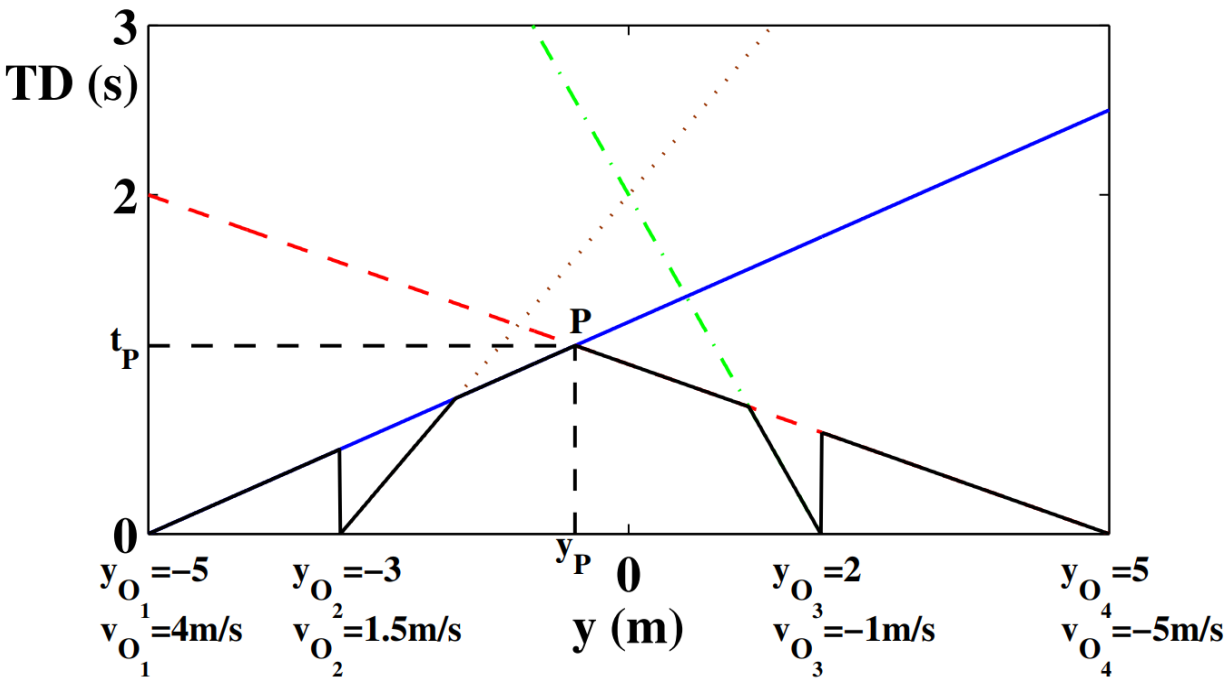

**Fig. 1.** Up: Four mobile points $O_1$-$O_4$ in a one dimensional space (axis y). Down: Diagram of the TD profile of $O_1$-$O_4$ with respect to the axis y (solid blue, dotted brown, dash-dot green, and dashed red) and intersection of them (solid black) which illustrates the TD of the set of the 4 mobile points $O_1$-$O_4$ with respect to the axis y. $y_P$ is the safest location of the axis y because its TD with respect to the set of the 4 mobile points $O_1$-$O_4$ is maximum. Note that the axis y is limited to [-5,5].

Now, if we generalize the location of point A to any arbitrary point on the axis $y$, Eq. (8) is obtained. In this equation, $TD_{B/y}(y)$ is the TD of point B with respect to any location on the axis y (it means that how much time will be passed until point B reaches any location on the axis $y$, with the assumption that $\upsilon_{B/y}$ is constant during the whole period). Plotting the diagram of Eq. (8) in a space-time coordinate reference with the horizontal axis as $y$, and the vertical axis as TD is a helpful representation of the TD of point B with respect to different points of the axis $y$. Figure 1 illustrates the diagram of the TD profiles of 4 mobile points $O_1 - O_4$ (moving on the axis $y$) with respect to the axis $y$. For each of the points $O_1 - O_4$, the TD profile is a line with constant slope $\left(\upsilon_{O_i/y}\right)^{-1}$ which starts from zero at the point $y_{O_i}$ and never possesses negative values. The black profile in the diagram is the TD profile of the set of all 4 mobile points $O_1 - O_4$ with respect to the axis $y$; that is the time remaining for any location on the axis y to be occupied by any mobile point. The approach to obtain the TD profile of a set of mobile points with respect to a space is to intersect between the TD profiles of all mobile points. Intersecting between TD profiles is similar to that between fuzzy sets by the "minimum" T-Norm. If G is a set of n mobile points (Eq. (9)), the TD of this set with respect to any location on the axis y is defined by Eq. (10). Furthermore, considering $O_1 - O_4$ as obstacles, the point $P$ is the safest point on the TD profile of the axis y with respect to the obstacle set G, and $y_P$ is the safest location on the axis $y$. This is because $TD_{G/y}(y_P)$ is the maximum value of $TD_{G/y}(y)$. Determining P and $y_P$ is the basis for path planning in later sections. The approach to determine $y_P$ is also similar to the "height method" of defuzzification. $t_P$ is the TD of the point $y_P$ with respect to the obstacle set G. The expression to obtain $t_P$ is again similar to the "maximum of minimum" fuzzy inference method (Eq. (11)). In order to obtain $y_P$, $t_P$ must be obtained first. Since $t_P$ is the value of $TD_{G/y}(y_P)$, the operator $TD^{-1}$ is defined to obtain $y_P$ (Eq. (12)).

$$TD_{B/y}(y) = 2\left(\operatorname{sign}\left(\frac{y - y_B}{\upsilon_{B/y}}\right) + 1\right)^{-1}\left|\frac{y - y_B}{\upsilon_{B/y}}\right| \tag{8}$$

$$G = \left\{O_k \middle| k = 1, \ldots, n\right\} \tag{9}$$

$$TD_{G/y}(y) = \min_{k=1}^{n}\left(TD_{O_k/y}(y)\right) \tag{10}$$

$$t_P = \max_{y=-5}^{5}\left(TD_{G/y}(y)\right) = \max_{y=-5}^{5}\left(\min_{k=1}^{n}\left(TD_{O_k/y}(y)\right)\right) \tag{11}$$

$$y_P = TD^{-1}(t_P) = TD^{-1}\left(\max_{y=-5}^{5}\left(\min_{k=1}^{n}\left(TD_{O_k/y}(y)\right)\right)\right) \tag{12}$$

### 2.2 Two dimensional space

To derive TD functions for the two dimensional space, consider point B moving in the direction of axis $x_1$ in a two dimensional space with coordinate reference $x - y$ (Fig. 2). If we use Eq. (8) to express the TD of point B with respect to this two dimensional space, Eq. (13) is obtained. However, since Eq. (8) was obtained for the one dimensional space, Eq. (13) only works for the points that are on the path of travel of point B. Figures 3 and 4 illustrate the diagram of Eq. (13) and its contour plot in the $xy$ plane, respectively. The whole points on these two diagrams do not show the TD of point B with respect to the $xy$ plane. In Fig. 3, only the points on top of point B's path of travel belong to the TD diagram of point B. Similarly, in Fig. 4, only the points on the intersection of contour lines and point B's path of travel show the TD of point B correctly. Therefore, Eq. (13) must be modified to be compatible with the definition of TD. In other words, the value of the TD of point B with respect to the points that are not on point B's path of travel must be $+\infty$. However, since our purpose is to construct a path planning framework which makes the vehicle avoid colliding with obstacles, obtaining the TD of a point with respect to a two dimensional space is not beneficial. This is because obstacles are not points in practice. Obstacles can be modeled by polygons, circles, etc. Accordingly, we do not proceed by modifying Eq. (13). Instead, we derive TD functions for line segments, polygons, and circles.

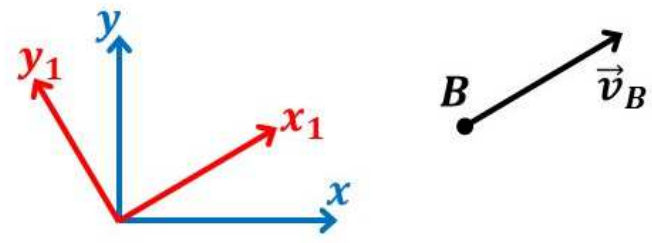

**Fig. 2.** Point B moving in a two dimensional space. The coordinate reference $x_1$-$y_1$ is chosen so that the axis $x_1$ is in the direction of the velocity vector of point B.

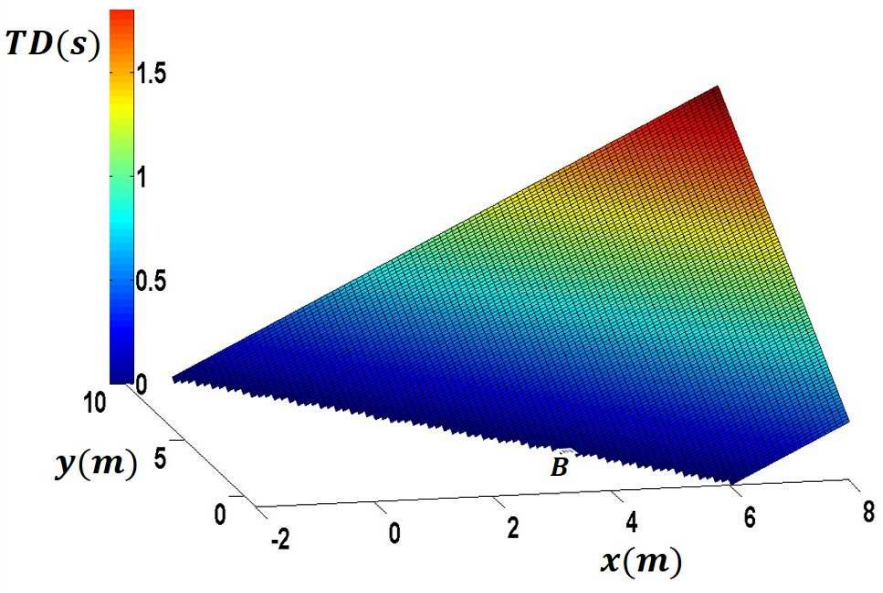

**Fig. 3.** Diagram of the TD of point B calculated by Eq. (13). Obviously, TD diagram of a point must be a line, not a surface. Therefore, Eq. (13) must be modified in order to illustrate the TD profile of a point with respect to a two dimensional space.

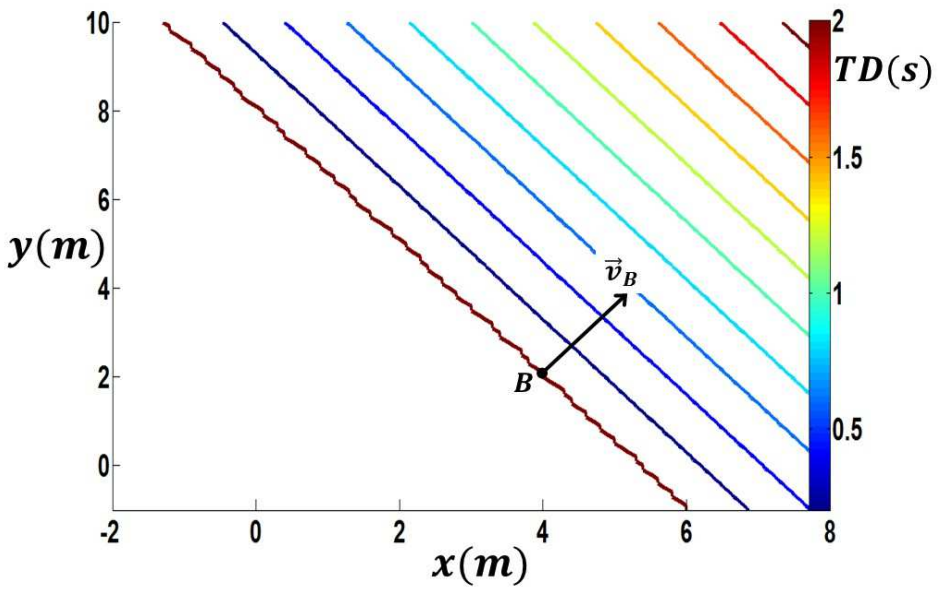

**Fig. 4.** Contour plot of the diagram of the TD of point B calculated by Eq. (13). Only the points that are on the intersection of point B's path of travel and the contour lines illustrate the TD of this point with respect to the two dimensional space.

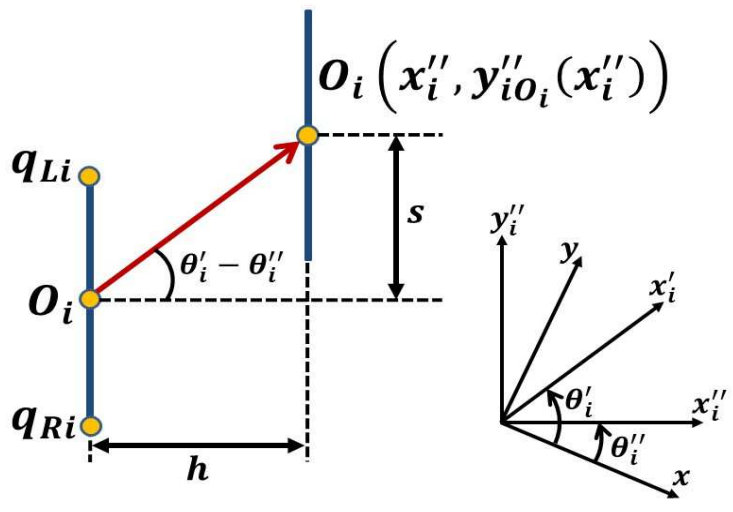

**Fig. 5.** Line segment $q_{R_i}q_{L_i}$ with center point $O_i$ moving in the positive direction of axis $x_i'$ with axis $x_i''$ normal to it.

$$TD_{B/xy}(x,y) = 2\left(\text{sign}\left(\frac{x_1 - x_{1B}}{v_{B/xy}}\right)+1\right)^{-1}\left|\frac{x_1 - x_{1B}}{v_{B/xy}}\right| \tag{13}$$

The contour lines of Fig. 4 can illustrate the TD of the edges of polygons if their lengths are restricted. Figure 5 illustrates a line segment ($q_{R_i}q_{L_i}$) with length $l_i$ and center point $O_i$ moving in the positive direction of axis $x_i'$ with axis $x_i''$ normal to it. Equations (14 - 17) indicate the relation between the coordinate reference $x-y$ and these two coordinates. In Eq. (15), $v_{i_y/xy}$ and $v_{i_x/xy}$ are the velocity components of point $O_i$ with respect to the $xy$ plane, in the $y$ and $x$ directions, respectively. Equations (18) and (19) denote the components of point $O_i$ 's initial location ($x^0_{O_i}$ and $y^0_{O_i}$), and Eq. (20) determines the length of the line segment $q_{R_i}q_{L_i}$.

$$\begin{bmatrix} x_i' \\ y_i' \end{bmatrix} = \begin{bmatrix} \cos\theta_i' & \sin\theta_i' \\ -\sin\theta_i' & \cos\theta_i' \end{bmatrix}\begin{bmatrix} x \\ y \end{bmatrix} \tag{14}$$

$$\theta_i' = \tan^{-1}\left(\frac{\upsilon_{i_y/xy}}{\upsilon_{i_x/xy}}\right) - \frac{\pi}{2}\left(\operatorname{sign}\left(\upsilon_{i_x/xy}\right) - 1\right)\operatorname{sign}\left(\upsilon_{i_x/xy}\right) \tag{15}$$

$$\begin{bmatrix} x_i'' \\ y_i'' \end{bmatrix} = \begin{bmatrix} \cos\theta_i'' & \sin\theta_i'' \\ -\sin\theta_i'' & \cos\theta_i'' \end{bmatrix} \begin{bmatrix} x \\ y \end{bmatrix} \tag{16}$$

$$\theta_i'' = \tan^{-1}\left(\frac{y_{q_{Ri}} - y_{q_{Li}}}{x_{q_{Ri}} - x_{q_{Li}}}\right) + \frac{\pi}{2} \tag{17}$$

$$x_{O_i}^0 = \frac{x_{q_{Ri}} + x_{q_{Li}}}{2} \tag{18}$$

$$y_{O_i}^0 = \frac{y_{q_{Ri}} + y_{q_{Li}}}{2} \tag{19}$$

$$l_i = \sqrt{\left(x_{q_{Ri}} - x_{q_{Li}}\right)^2 + \left(y_{q_{Ri}} - y_{q_{Li}}\right)^2} \tag{20}$$

The contour lines of the TD diagram of the line segment $q_{R_i}q_{L_i}$ are a set of parallel line segments with length $l_i$ and center point $O_i(x_i'', y_{iO_i}''(x_i''))$ perpendicular to the $x_i''$ axis. Note that since the point $O_i$ moves in the definite direction of the axis $x_i'$, the components $x_i''$ and $y_i''$ of this point depend on each other. In other words, $y_{iO_i}''(x_i'')$ is a function that determines $y_{iO_i}''$ for every value of $x_i''$. To obtain this function, according to Fig. 5 one has:

$$y_{iO_i}''\left(x_i''\right) = y_{iO_i}''^{0} + s$$

$$s = h\tan(\theta_i' - \theta_i'')$$

$$h = x_i'' - x_{iO_i}''^{0}$$

Hence:

$$y_{iO_i}''\left(x_i''\right) = y_{iO_i}''^{0} + \left(x_i'' - x_{iO_i}''^{0}\right)\tan(\theta_i' - \theta_i'') \tag{21}$$

where $x_{iO_i}''^{0}$ and $y_{iO_i}''^{0}$ are the components of the initial location of point $O_i$. Using this function, the set of the mentioned parallel line segments with length $l_i$ and center point $O_i(x_i'', y_{iO_i}''(x_i''))$ perpendicular to the axis $x_i''$ is defined by Eq. (22). Consequently, the TD of the line segment $q_{R_i}q_{L_i}$ is represented by Eq. (23). In this equation, $\upsilon_{x_i'O_i/xy}$ is the velocity of point $O_i$ with respect to the $xy$ plane, in the $x_i''$ direction ($\upsilon_{x_i''O_i/xy} = \upsilon_{i_x/xy}\cos\theta_i'' + \upsilon_{i_y/xy}\sin\theta_i''$).

$$e_i = \left\{(x_i'', y_i'') \middle| \left|y_i'' - y_{iO_i}''(x_i'')\right| \le \frac{l_i}{2}\right\} \tag{22}$$

$$TD_{q_{L_i}q_{R_i}/xy}(x,y) = \begin{cases} 2\left(\operatorname{sign}\left(\dfrac{x_i'' - x_{iO_i}''^{0}}{\upsilon_{x_i'O_i/xy}}\right) + 1\right)^{-1}\left|\dfrac{x_i'' - x_{iO_i}''^{0}}{\upsilon_{x_i'O_i/xy}}\right|, & (x_i'', y_i'') \in e_i \\ +\infty, & (x_i'', y_i'') \notin e_i \end{cases} \tag{23}$$

To integrate Eq. (22) and Eq. (23) in order to obtain an explicit equation for the function $TD_{q_{L_i}q_{R_i}/xy}(x,y)$, the multiplier $Q_i$ is defined with the following feature:

$$\begin{cases} Q_i = 1 \Leftrightarrow (x_i'', y_i'') \in e_i \\ Q_i = 0 \Leftrightarrow (x_i'', y_i'') \notin e_i \end{cases} \tag{24}$$

An expression for $Q_i$ is:

$$Q_i = \operatorname{sign}\left(\operatorname{sign}\left(\frac{l_i}{2} - \left|y_i'' - y_{iO_i}''(x_i'')\right|\right) + 1\right) \tag{25}$$

Representing the line segment $q_{R_i}q_{L_i}$ with its center point $O_i$ (for conciseness), finally, the TD of the line segment $O_i$ with respect to the $xy$ plane is expressed by Eq. (26):

$$TD_{O_i/xy}(x,y) = 2Q_i^{-1}\left(\operatorname{sign}\left(\frac{x_i'' - x_{iO_i}''^{0}}{\upsilon_{x_i''O_i/xy}}\right) + 1\right)^{-1}\left|\frac{x_i'' - x_{iO_i}''^{0}}{\upsilon_{x_i''O_i/xy}}\right| \tag{26}$$

Defining a polygon as a set of line segments (edges), the TD of every edge of the polygon is obtained from Eq. (26), and similar to Eq. (10), Eq. (27) expresses the TD of the polygon (with $m$ edges) with respect to the $xy$ plane. Note that the $x_i''$ axis of every edge of the polygon is perpendicular to the edge. As a result, $x_i''$ axes of parallel edges can be the same. (As it is explained in the next section, for Z-Infinity functions, $x_i''$ axes of parallel edges cannot be the same.)

$$TD_{Ply/xy}(x,y)=\min_{i=1}^{m}\left(TD_{O_i/xy}(x,y)\right) \tag{27}$$

Similar to the one dimensional case in which Eq. (8) and Eq. (10) were used to plot the TD profile of points and a set of points, Eq. (26) and Eq. (27) can be used to plot the TD surface of line segments and polygons. Figures 6 and 7 illustrate the TD surface of a rectangle and a triangle, and its contour plot, with respect to *xy* plane, respectively. The TD surface of Fig. 6 also represents outermost boundaries of swept volumes for these two polygons. Note that the value of the TD surface inside the borders of an object's current location must be zero. TD functions do not have the capability to make the value of the TD surface in such areas equal to zero. Accordingly, Z-Infinity functions are defined in the next section to make the value of the TD surface of objects in these areas equal to zero.

Similar to polygons, the TD of a circle with center point $O_j$ and radius $R_j$ moving in the positive direction of an axis $x'_j$ (Fig. 8), with respect to the *xy* plane, is represented by Eq. (28):

$$TD_{O_j/xy}(x,y)=2Q_j^{-1}\left(\mathrm{sign}\left(\frac{x'_j-x'^{0}_{jO_j}-Q_j\sqrt{R_j^2-(y'_j-y'^{0}_{jO_j})^2}}{v_{O_j/xy}}\right)+1\right)^{-1}\times\left|\frac{x'_j-x'^{0}_{jO_j}-Q_j\sqrt{R_j^2-(y'_j-y'^{0}_{jO_j})^2}}{v_{O_j/xy}}\right| \tag{28}$$

In this equation, $Q_j$ can be one of the following expressions:

$$Q_j=1-\left|\mathrm{sign}\left(\mathrm{Im}\left(\sqrt{R_j^2-(y'_j-y'^{0}_{jO_j})^2}\right)\right)\right| \tag{29}$$

Or:

$$Q_j=\mathrm{sign}\left(\mathrm{sign}\left(R_j-\left|y'_j-y'^{0}_{jO_j}\right|\right)+1\right) \tag{30}$$

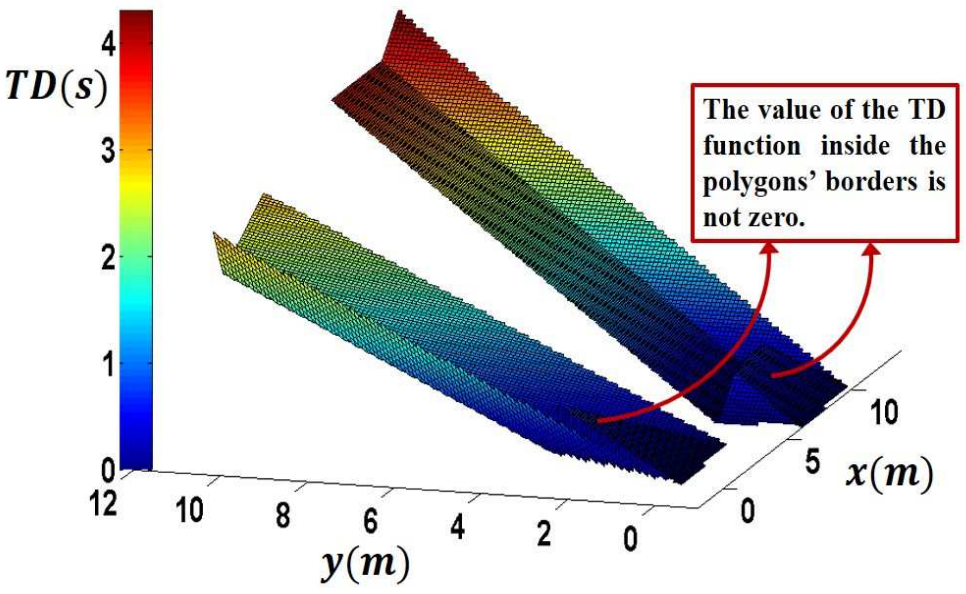

**Fig. 6.** Diagram of the TD surface of a rectangle and a triangle. This diagram can also be interpreted as the diagram of outermost boundaries of swept volumes. Logically, the value of the TD surface inside the borders of objects' current location must be zero. Z-Infinity functions are defined in Section 3 for this purpose.

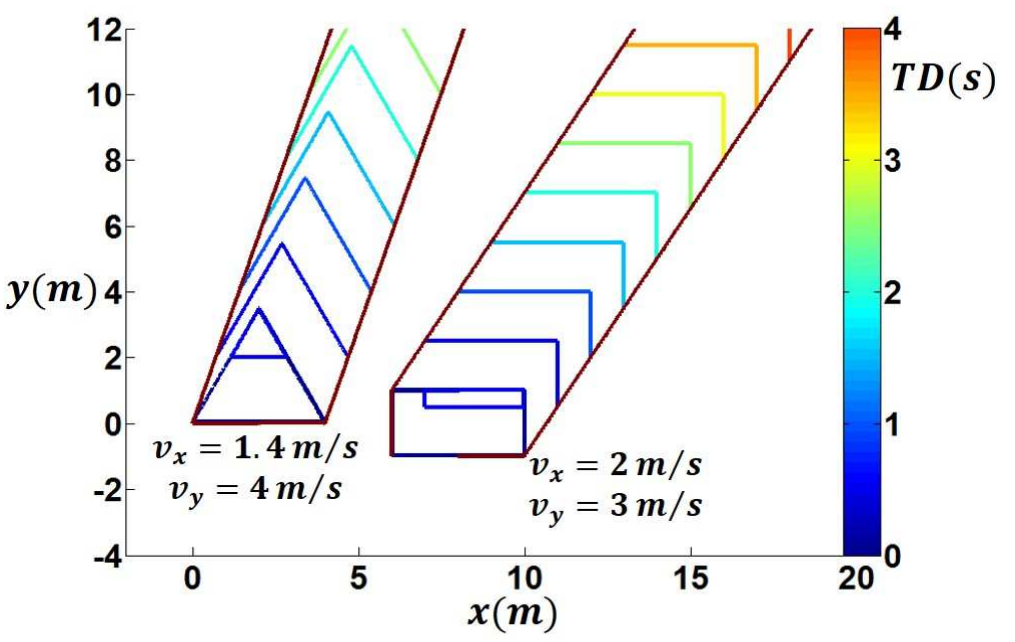

**Fig. 7.** Contour plot of the TD surface of Fig. 6.

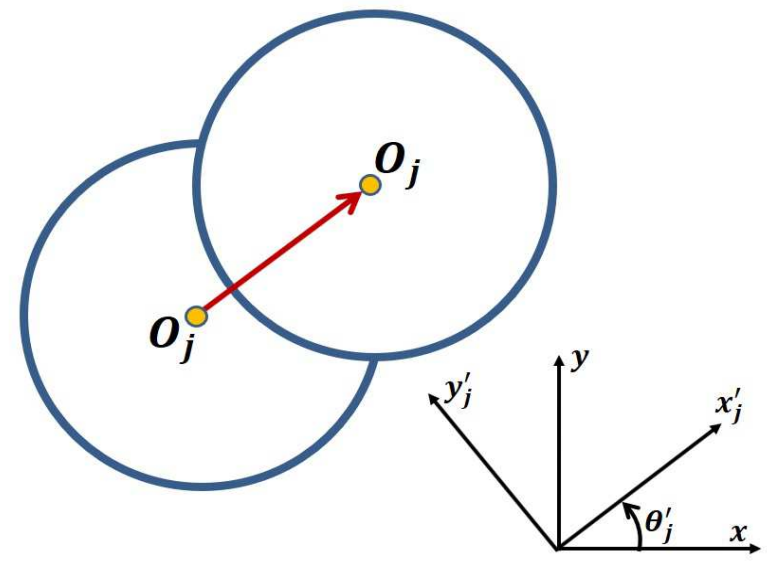

**Fig. 8.** A circle moving in a two dimensional space.

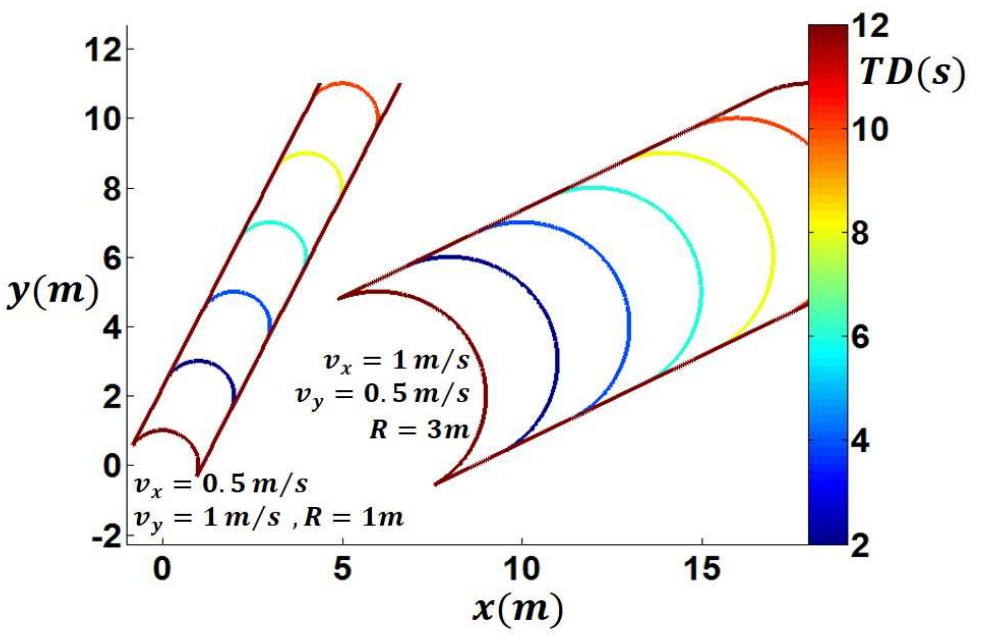

**Fig. 9.** Contour plot of the TD surface of two circles.

In Eq. (29), $\mathrm{Im}(\sqrt{R_j^2 - (y'_j - y'^0_{jO_j})^2})$ returns the imaginary part of $\sqrt{R_j^2 - (y'_j - y'^0_{jO_j})^2}$ . $x'^0_{jO_j}$ and $y'^0_{jO_j}$ (in Eq. (28)) are the components of the initial location of point $O_j$ , and $v_{O_j/xy}$ is the magnitude of the circle's velocity with respect to the $xy$ plane (Eq. (31)). Substituting $i$ with $j$ in Eq. (14) and Eq. (15), the relation between the coordinate references $x - y$ and $x'_j - y'_j$ is obtained. Figure 9 demonstrates the contour plot of the TD surface of two circles with respect to the $xy$ plane.

$$v_{O_j/xy} = \sqrt{v^2_{j_x/xy} + v^2_{j_y/xy}} \tag{31}$$

## 3. Path Planning

In this section, the details of the path planning framework are explained. In this framework, the vehicle is considered as a point robot in the configuration space, and the obstacles are, therefore, enlarged to turn into configuration obstacles. Then, the TD surface of the configuration space is calculated, and the route is obtained in the configuration-time space. The configuration-time space is constructed by embedding TD (the time dimension) to the configuration space. The projection of the route onto the configuration space, which is obtained by the operator $TD^{-1}$, is the planned path which is followed by the vehicle.

### 3.1 Z-Infinity Functions

As mentioned in Section 2, the value of the TD surface inside the borders of objects' current location must be zero. Therefore, Zero-Infinity (Z-Infinity or $Z_\infty$ ) functions are defined in this section.

Similar to TD functions, the $Z_\infty$ function for a polygon is composed of the $Z_\infty$ functions of the polygon's edges. The $Z_\infty$ function of an edge is zero on one of its sides and infinity on the other. Equation (32) expresses the $Z_\infty$ function for the line segment of Fig. 5:

$$Z_{\infty_{O_i}}(x,y) = \left(\mathrm{sign}\left(\mathrm{sign}\left(x''^0_{iO_i} - x''_i\right) + 1\right)\right)^{-1} - 1 \tag{32}$$

where $x''^0_{iO_i}$ and $x''_i$ are already defined in Section 2. However, in contrast to TD functions, the direction of $x''_i$ for $Z_\infty$ differs for parallel edges of a polygon; its positive direction is pointing outwards the polygon (this definition does not interrupt TD functions). Therefore, $\theta''_i$ is redefined. The vertex points $q_{R_i}$ and $q_{L_i}$ of every edge are defined so that if we stand on the point $O_i$ facing toward the positive direction of the $x''_i$ axis, the point $q_{R_i}$ will be located at our right side and $q_{L_i}$ at the left side. Consequently, $\theta''_i$ of every edge is expressed by Eq. (33):

$$\theta''_i = \tan^{-1}\left(\frac{y_{q_{Ri}} - y_{q_{Li}}}{x_{q_{Ri}} - x_{q_{Li}}}\right) + \frac{\pi}{2}\mathrm{sign}\left(\mathrm{sign}\left(x_{q_{Ri}} - x_{q_{Li}}\right) + 0.5\right) \tag{33}$$

Finally, Eq. (34) expresses the $Z_\infty$ function for a convex polygon with $m$ edges:

$$Z_{\infty_{Ply}}(x,y) = \max_{i=1}^{m}\left(Z_{\infty_{O_i}}(x,y)\right) \tag{34}$$

For rectangular and circular shape obstacles there are simpler $Z_\infty$ functions. Equations (35) and (36) express the $Z_\infty$ function for a rectangle and a circle, respectively:

$$Z_{\infty_{Rec}} = \left(\mathrm{sign}\left(\left(\mathrm{sign}\left(\frac{l}{2} - \left|x''_w - x''_{w,C}\right|\right) + 1\right) \times \left(\mathrm{sign}\left(\frac{w}{2} - \left|x''_l - x''_{l,C}\right|\right) + 1\right)\right)\right)^{-1} - 1 \tag{35}$$

$$Z_{\infty_{Crcl}} = \left(\mathrm{sign}\left(\mathrm{sign}\left(R^2 - (x - x_C)^2 - (y - y_C)^2\right) + 1\right)\right)^{-1} - 1 \tag{36}$$

In Eq. (35), $w$ and $l$ are the rectangle's width and length, and $x''_w$ and $x''_l$ are axes perpendicular to them, respectively. $x''_{w,C}$ and $x''_{l,C}$ are the components of the location of the rectangle's center on the mentioned axes. In Eq. (36), $x_C$ and $y_C$ are the components of the circle's center, and $R$ is the circle's radius.

### 3.2 Required Coordinate References

In this section, three coordinate references, which are used in the structure of the framework, are defined.

The Global Coordinate reference (GC) $X - Y$ is a stationary coordinate reference with no preferred specific directions for its axes. The work space is constructed on this coordinate reference.

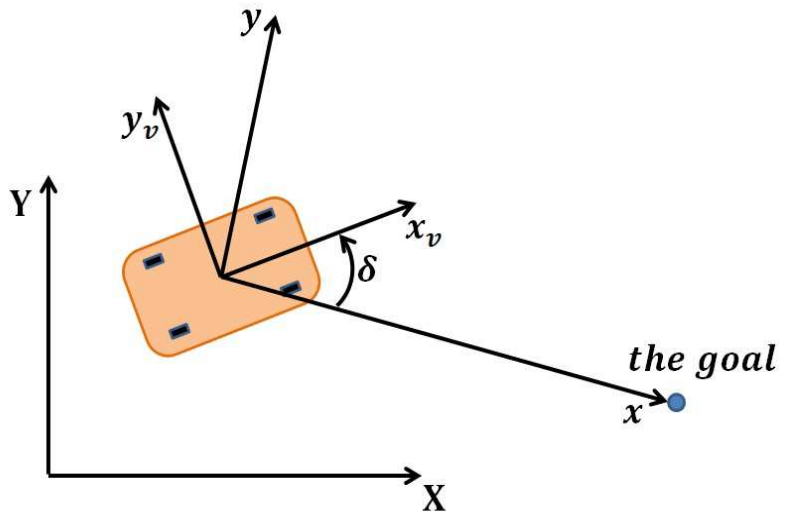

**Fig. 10.** Illustration of GC, PC and VC.

The Vehicle's Coordinate reference (VC) $x_v - y_v$ is a coordinate reference with its origin at the vehicle's reference point, the $x_v$ axis toward the vehicle's heading and normal to the vehicle's front edge, and the $y_v$ axis normal to the left edge.

The Principal Coordinate reference (PC) $x - y$ is a coordinate reference with its origin at the vehicle's reference point, the $x$ axis pointing from the vehicle's reference point to the goal, and the $y$ axis perpendicular to the axis $x$ in the plane of motion. This coordinate reference is a rotated coordinate reference with respect to VC.

PC and VC are reconstructed in every sequence of path planning. Figure 10 illustrates the three coordinate references. The configuration space is constructed on the PC. Therefore, the configuration space is the $xy$ plane which its velocity is equal to the vehicle's velocity.

### 3.3 The Path's Equation

To obtain the collision free path from the vehicle's reference point to the goal, first, the TD surface of the configuration space with respect to the configuration obstacles, $TD_{conf}$ , must be calculated. To carry this out, Eq. (10) is generalized to the two dimensional space. As a result, $TD_{conf}$ is obtained from Eq. (37). In this equation, the term $TD_{O_k/xy}(x,y)$ gives the TD of the configuration obstacles with respect to the configuration space and is obtained from Eq. (26) or Eq. (28) depending on the geometry of every obstacle. Note that Z-Infinity functions are also augmented to this equation, and there is a $Z_\infty$ function for every configuration obstacle.

$$TD_{conf} = \min_{k=1}^{n}\left(TD_{O_k/xy}(x,y); Z_{\infty_{obs_1}}(x,y), Z_{\infty_{obs_2}}(x,y), \ldots\right) \tag{37}$$

As mentioned previously, determining the safest point $P$ on the TD profile and the safest location $y_P$ on the axis y in Fig. 1 is the basis of path planning. In Fig. 1, the safest point on the TD profile is a point the TD of which is maximum with respect to the obstacle set G. Considering part of the configuration-time space between the robot and the goal to be sectioned by planes perpendicular to the axis $x$ , for every value of $x$ there is a TD profile on the intersection of the plane crossing the section $x$ and $TD_{conf}$. This TD profile is similar to the TD profile of Fig. 1 on which there is a safest point. The route is formed by connecting these safest points consecutively. Then, the path is obtained as the projection of the route onto the configuration space using the operator $TD^{-1}$. However, since the safest point in a section is not guaranteed to be safe for the vehicle, the parameter $T_s$ is defined. If the TD of the safest point in a section is equal or larger than $T_s$ , the safest point is also safe for the vehicle; otherwise, the last safe point will be the end point of the route ( $T_s$ can be calculated as a function of the parameters of the road and the vehicle). As a result, the path is obtained from Eq. (38) which is the generalized form of Eq. (12). In this equation, $t_P(x)$ is the TD of the safest point in section $x$ which is obtained from Eq. (39). Note that Eq. (39) is also the generalized form of Eq. (11) and $y_{min}$ and $y_{max}$ in this equation are the minimum and maximum values of y in the configuration space.

$$y_P(x) = TD^{-1}\left(t_P(x)\right) \quad , \quad t_P(x) \geq T_s \tag{38}$$

$$t_P(x) = \max_{y_{min}}^{y_{max}}\left(TD_{conf}\right) \tag{39}$$

### 3.4 Route Function

Route Function (RF) plays a leading role in path planning. RF guides the vehicle toward the goal and provides length optimality in the path.

As explained previously, the route is formed by connecting the safest points of consecutive sections of the configuration-time space. However, the safest points may be distributed disorderly in the configuration-time space. Therefore, by connecting the safest points and projecting them to the configuration space, a jagged path is obtained which does not provide length optimality. To resolve this problem, the safest points are manipulated by the RF.

To derive an expression for the RF, and clarify how this function manipulates the safest points, we first define the set of the safe points in the configuration-time space. The set of the safe points ( $S_s$ ) is defined as the points on $TD_{conf}$ which their TD is equal or greater than $T_s$ . The RF's duty is to facilitate generating a route through $S_s$ from the vehicle's reference point to the goal, considering length optimality. In order to provide length optimality, every point of the route must be the optimal point in its corresponding section $x$ . Ideally, the optimal point in a section $x$ is a point belonging to $S_s$ which has the best location for providing length optimality in the whole path. However, for the sake of quickness, we prefer to define the RF so that the optimal point in every section $x$ is obtained independent of other sections. Although this strategy has a negative impact on length optimality, it can result in a faster algorithm because the optimal points of all sections can be found simultaneously. Accordingly, we define the cost function for our optimization problem as $J = |y|$ in every section $x$ ; therefore:

$$J(x) = |y| \tag{40}$$

As a result, in every section $x$ , the point with minimum $|y|$ in $S_s$ is the optimal (or near optimal) point. Therefore, the path will get close to the axis $x$ as possible.

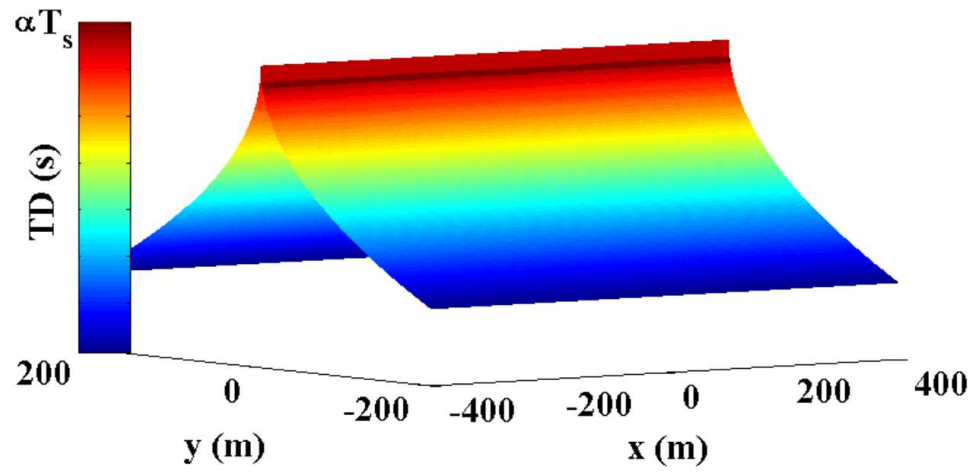

**Fig. 11.** Diagram of the surface of RF expressed by Eq. (41).

Although calculating the $|y|$ for every member of $S_s$ and finding the point with minimum $|y|$ in every section $x$ is a simple and straight solution for this optimization problem, it does not result in a fast algorithm. Accordingly, we solve this optimization problem such that $S_s$ is not calculated in practice, and a faster algorithm is achieved.

Equation (41) presents an expression for the RF. In this function, $\alpha$, $\beta$, and $\gamma$ are constant positive values ($\beta, \gamma \leq 1 < \alpha$). Diagram of this function is illustrated in Fig. 11. This surface is a little bit greater than $T_s$ and increases very slightly as it gets close to the axis $x$. If $TD_{conf}$ becomes similar to the surface of the RF, when it increases (in every section $x$), the performance index J decreases. Accordingly, J is minimized in the safest point of every section $x$. In other words, by finding the maximum of $TD_{conf}$ (in every section $x$), which is done in Eq. (39), both the optimization problem and the collision avoidance problem are solved.

$$RF(x,y) = \alpha T_s - \beta |y|^{\gamma} \tag{41}$$

To make $TD_{conf}$ similar to the RF, the RF is augmented to Eq. (37); therefore, $TD_{conf}$ is obtained from Eq. (42):

$$TD_{conf} = \min_{k=1}^{n}\left(TD_{O_k/xy}(x,y); RF(x,y); Z_{\infty_{obs_1}}(x,y), Z_{\infty_{obs_2}}(x,y), \ldots\right) \tag{42}$$

Now, $TD_{conf}$ contains two kinds of points:

1. The points that were over the RF which are now excluded and replaced with part of the RF under them.

2. The points that were under the RF which have excluded part of the RF over them; these points are called the under RF points.

Therefore, the construction of $TD_{conf}$ is similar to the RF itself. That is the points with higher TD are closer to the axis $x$ (have lower $|y|$) except for the under RF points. The under RF points that do not belong to $S_s$ are not capable to be part of the route because they belong to the obstacles or collision risky areas. Some of the under RF points that belong to $S_s$ can be part of the route; however, for the sake of quickness, the algorithm only finds some of them although it may be in contrast with length optimization.

Finally, by substituting Eq. (42) in Eq. (39), $t_P(x)$ in Eq. (38) is obtained from Eq. (43). As a result, the collision avoidance problem and the length optimization problem are solved simultaneously by finding the maximum value of $TD_{conf}$ (the safest point) in every section $x$.

$$t_P(x) = \max_{y_{min}}^{y_{max}}\left(\min_{k=1}^{n}\left(TD_{O_k/xy}(x,y); RF(x,y); Z_{\infty_{obs1}}(x,y), Z_{\infty_{obs2}}(x,y), \ldots\right)\right) \tag{43}$$

The simple performance index defined for length optimization and ignoring some of the under RF points that belong to $S_s$, are two factors that can reduce the quality of length optimization in our formulation. However, the impact of the latter can be reduced by selecting appropriate values for the parameters $\alpha$, $\beta$, and $\gamma$. As mentioned before, the value of RF must be a little bit greater than $T_s$ and increase very slightly as it gets close to the axis $x$. The lower the value of RF and its gradient, the lower the number of under RF points that belong to $S_s$.

### 3.5 Jumping

According to the definition of PC, the axis $x$ of PC connects the vehicle's reference point to the goal. Therefore, the axis $x$ goes through obstacles and the safest point may jump from one side of an obstacle to its other side. We call this phenomenon "jumping". There are two situations in which jumping happens:

1. The jumping leads the vehicle to cross through an obstacle. For such situations, if the obstacle is narrow, jumping is more likely to happen and will be closer to the vehicle. Accordingly, to illustrate the worst jumping situations, the obstacles are considered as line segments with different orientations in Fig. 12.

2. The jumping is happened because the vehicle does not know which side of the obstacle is better for circumvention. Figure 13 illustrates an instance of this type of jumping.

To propose an strategy which can overcome the both types of jumping, first of all, the robot must have a look-ahead point to see variations in the path before reaching them. In addition, the path planning module must replan the path as the robot moves. The $x$ component of the look-ahead point is called the look-ahead distance of the vehicle. Equation (44) expresses the look-ahead distance $L$ as a function of $\delta$. In this equation, $\zeta$ is a positive multiplier, and $D$ is the vehicle's diameter. The $y$ component of the look-ahead point is $y_P(L)$. Therefore, the look-ahead point lies on the path at $x = L$.

$$L = \zeta D \times \max\left(\cos\delta, \frac{1}{2}\right) \tag{44}$$

Next, the RF must be modified. As it is illustrated in Fig. 11, the surface of RF is symmetric with respect to the axis $x$. When the path is likely to have a jumping, one side of the surface of RF must vanish. Therefore, there will be no path at one side of the axis $x$. As a result, jumping does not happen.

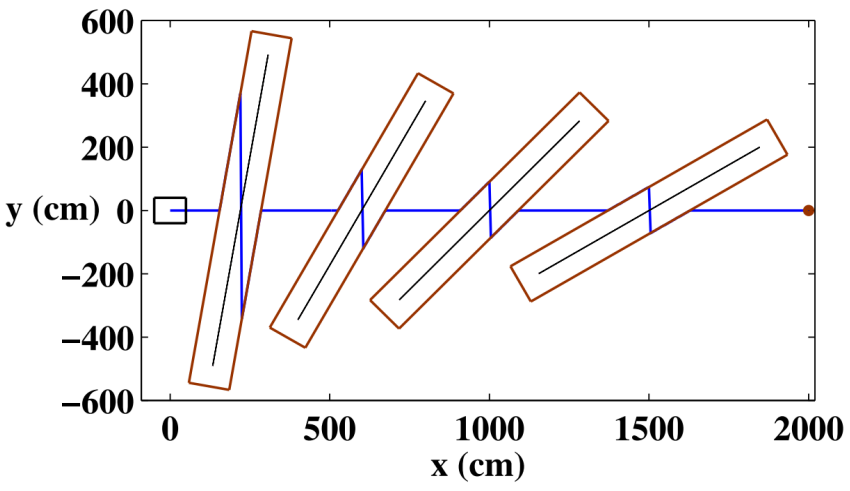

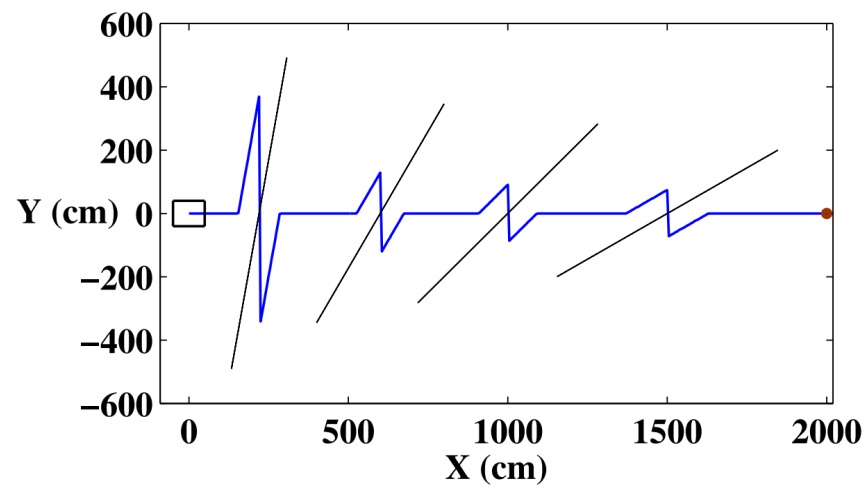

**Fig. 12.** Jumping type 1. Left: configuration space (the vehicle's borders are also illustrated in the configuration space). Right: work space.

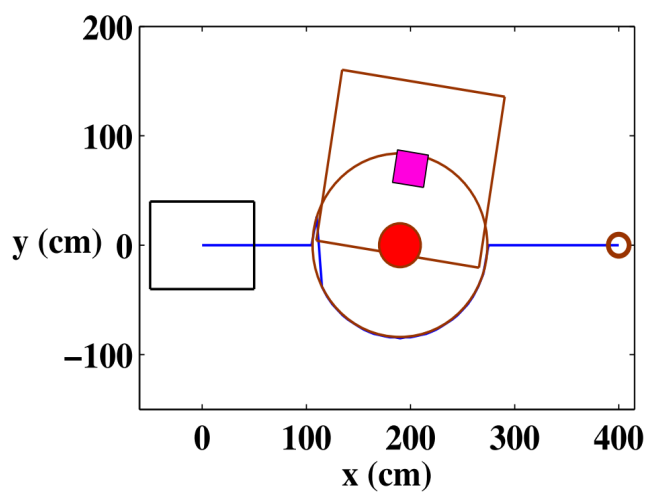

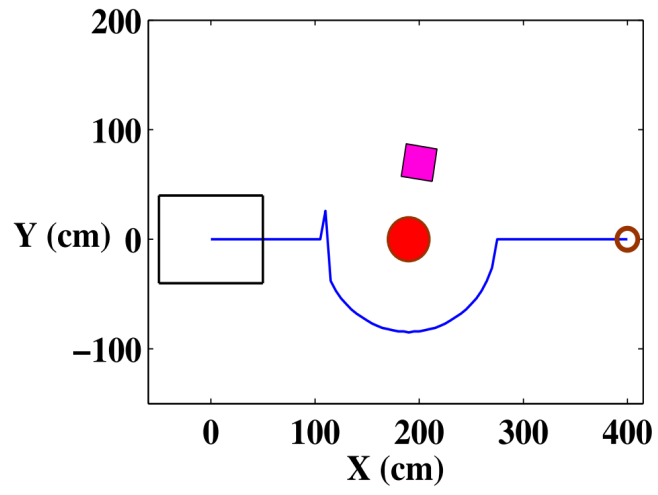

**Fig. 13.** Jumping type 2. Left: configuration space. Right: work space.

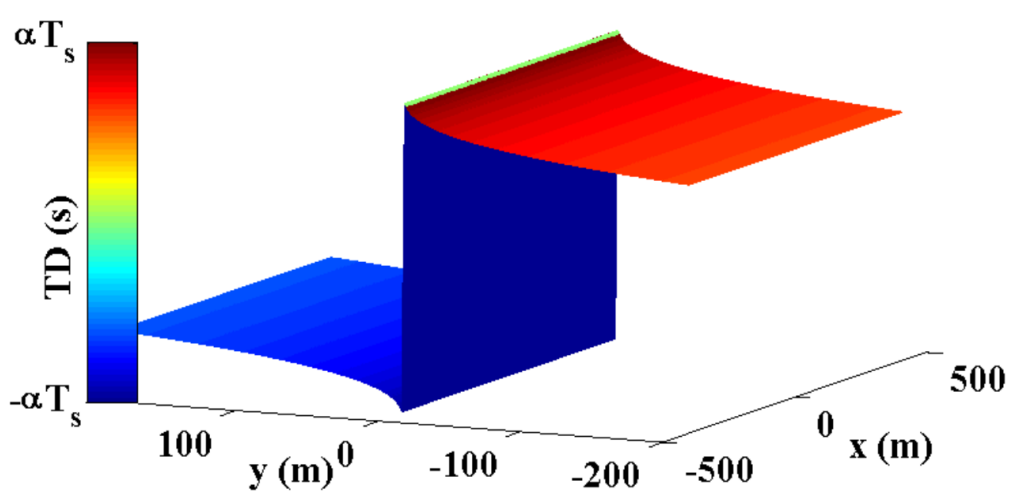

**Fig. 14.** Diagram of the surface of RF when $y_P(L) \le -\eta D$ . In this case, part of the RF belonging to the area with $y > 0$ has vanished (has got negative values).

In type 1 of jumping, when the path is faced with the obstacle, first it gets a considerable distance from the axis $x$ , and then jumping happens (long jumping). Therefore, if the path in the look-ahead point gets a considerable distance from the axis $x$ , it must not suddenly return toward the axis $x$ . Therefore, in such situations, subsequent parts of the path must be planned in the same side of the axis $x$ where the path started getting distance from the axis $x$ . Therefore, the value of RF over the other side of the axis $x$ must vanish until the path in the look-ahead point again gets close to the axis $x$ . Accordingly, RF is modified by Eq. (45). In this equation, D is the vehicle's diameter, $\eta > 0$ is a multiplier that determines how much distance from the axis $x$ is considerable, and $\lfloor . \rfloor$ is the floor function. Note that $y_P(L)$ is obtained from the last sequence of path planning. Figure 14 illustrates the surface of RF when $| y_P(L) | \ge \eta D$.

$$RF(x,y) = \left( \text{sign}\left( \text{sign}\left(y_P(L)\right) \left\lfloor \frac{|y_P(L)|}{\eta D} \right\rfloor \right) \text{sign}(y) + \left| \text{sign}\left( \left\lfloor \frac{|y_P(L)|}{\eta D} \right\rfloor \right) - 1 \right| \right) \left( \alpha T_s - \beta \, | y |^{\gamma} \right) \tag{45}$$

Type 2 of jumping contains situations that the path gets a small distance from the axis $x$ and returns toward it (small jumping). According to Eq. (45), when the look-ahead point reaches the point with small jumping, no side of RF vanishes. To resolve small jumpings, part of the path between the vehicle's reference point and the look-ahead point is replaced with a smooth trajectory. The vehicle always tracks this trajectory. This smooth trajectory also helps nonholonomic robots and ground vehicles to track the path. Figures 15 and 16 illustrate how the modified RF and the path smoothing module resolve the jumpings shown in Figs. 12 and 13.

In the subsequent sections, the framework is evaluated in simulations. The way this framework performs the path planning task can be compared with a real driver. A real driver plans a path from the vehicle's location to the destination in his/her mind. Part of the path which is close to the vehicle is accurate and considers collision avoidance. Accuracy is decreased in the rest of the path such that farther a part of the path to the vehicle, lower the accuracy of that part. Similarly, our framework generates a trajectory from the vehicle's reference point to the look-ahead point. This trajectory is the most accurate part of the path which is tracked by the vehicle. Accuracy in the rest of the path is decreased similar to a real driver's case. Therefore, parts of the path apart from the vehicle's look-ahead point may even contain collision with obstacles.

Since the path in every section $x$ is obtained independent of other sections, it is not necessary to calculate part of the path between the look-ahead point and the goal. This fact can decrease the computational cost greatly. Accordingly, in the structure of the Static and Dynamic TD methods, which are introduced in later sections, the functions $TD$ , $RF$ and $Z_\infty$ , which construct $TD_{conf}$ , and the path are only calculated in the interval $\{(x,y) \,|\, 0 < x < \zeta \times D \,,\, y_{min} < y < y_{max}\}$ . However, for illustration purposes we may calculate $TD_{conf}$ over the entire configuration space separately.

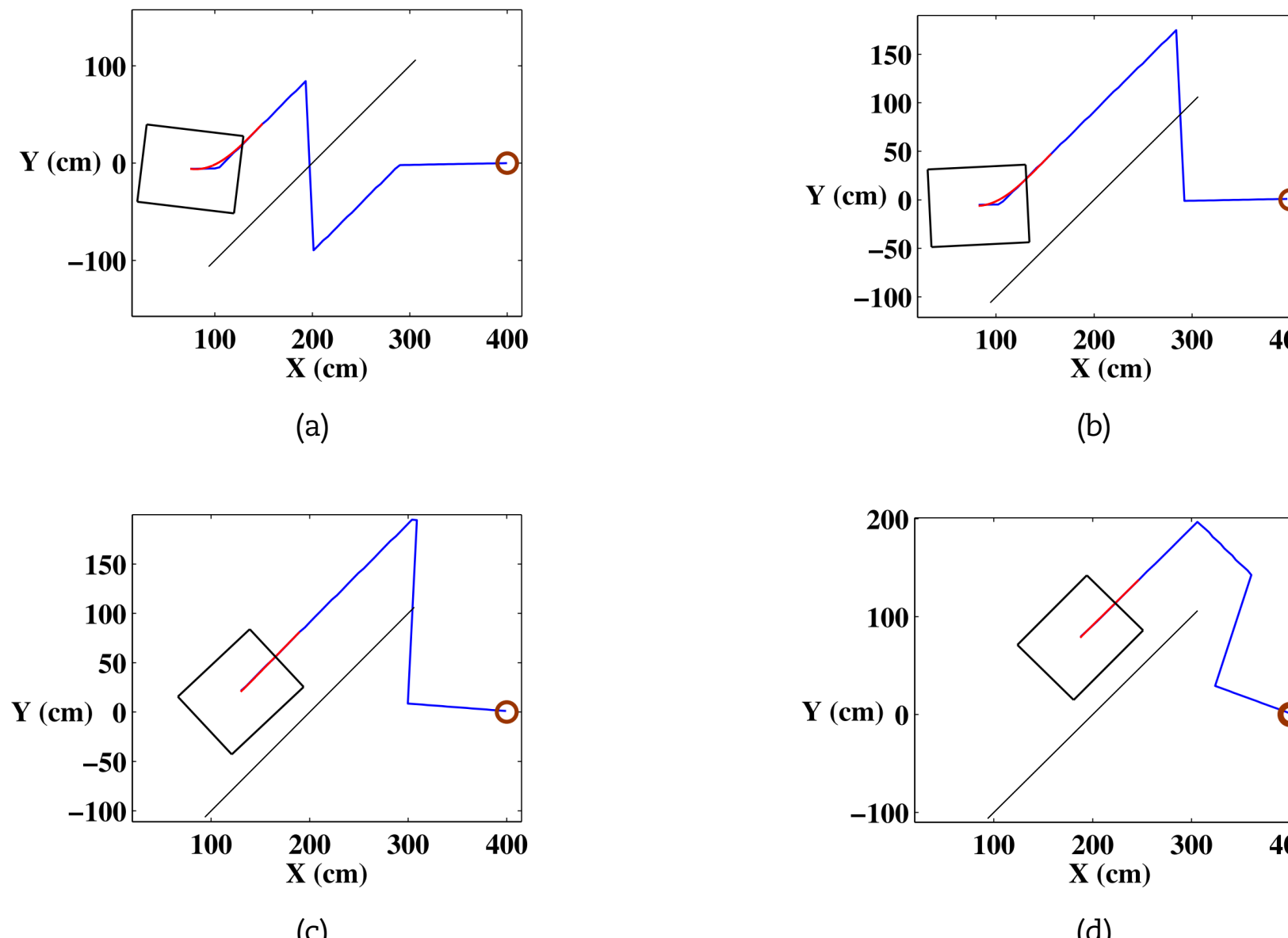

**Fig. 15.** The modified RF and replanning resolve jumpings of type 1.

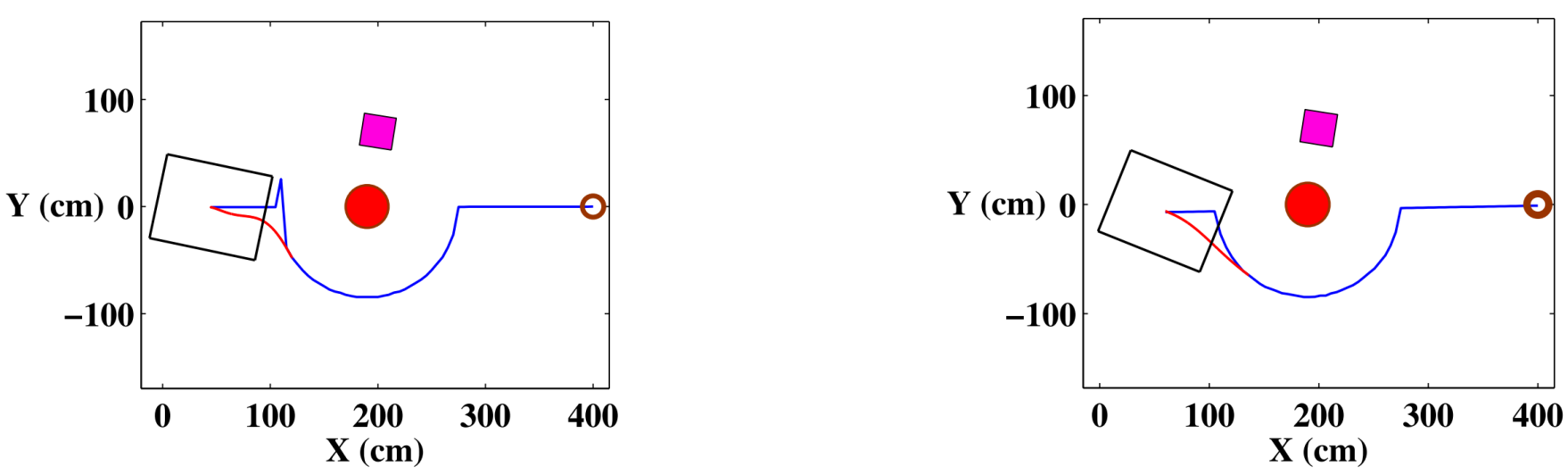

**Fig. 16.** The path smoothing module resolves jumpings of type 2.

### 3.6 The Static TD Method

For an environment with static obstacles, RF and $Z_\infty$ are sufficient to obtain $TD_{conf}$ , and it is not necessary to use TD functions. However, as it will be revealed later, using TD functions improves the optimality of the path.

By excluding TD functions from Eq. (43), and combining it with Eq. (38), Eq. (46) is obtained. Since TD functions are excluded from Eq. (43), the condition $t_P(x) \geq T_s$ in Eq. (38) is changed to $t_P(x) > 0$ in Eq. (46). We call this formulation, the "Static TD" method. It is worth emphasizing that RF is obtained from Eq. (45). The pseudo code of this algorithm is presented in Appendix 1.

$$y_P(x) = TD^{-1}\left(\max_{y_{min}}^{y_{max}}\left(\min\left(RF(x,y);Z_{\infty_{obs_1}}(x,y),Z_{\infty_{obs_2}}(x,y),\ldots\right)\right)\right),\max_{y_{min}}^{y_{max}}\left(\min\left(RF(x,y);Z_{\infty_{obs_1}}(x,y),Z_{\infty_{obs_2}}(x,y),\ldots\right)\right) > 0 \qquad (46)$$

#### 3.6.1 Simulation results

To evaluate the Static TD method and compare its performance with some popular algorithms, the path planning scenario of Fig. 17 with two stationary obstacles is considered. The size of the robot is $100mm \times 80mm$ and the geometric center of the vehicle is considered as the vehicle's reference point. Length of the edges of the polygon and the diameter of the circle are extended by the vehicle's diameter to obtain configuration obstacles. The path smoothing module generates a 5-degree polynomial curve from the vehicle's reference point to the look-ahead point using the SCR-Normalize method [40]. When 10% of the smooth trajectory is tracked by the vehicle, the path is replanned. Provided that the vehicle can exactly track the smooth trajectory, its motion is simulated. The vehicle's tracked trajectory from the start point to the goal is illustrated in Fig. 17. In addition, diagram of $TD_{conf}$ (over the whole configuration space) at the first moment of this scenario is illustrated in Fig. 18. Furthermore, some sequences of path planning are illustrated in Fig. 19. Table 1 indicates the values of the parameters of the Static TD's formulation for this example.

This path planning scenario is also solved by the algorithms RRT [41], Bidirectional RRT (BRRT) [42], PRM [43] and A* [44] which are visible in Fig. 17. The comparison indicates that the Static TD method is considerably faster than the other algorithms, and the path obtained from the static TD method is the shortest path after the paths obtained from the A* and PRM algorithms. Note that replanning is done for all of the algorithms except A* for which replanning does not have a considerable effect. The computer used for simulations contains an Intel Core™ i5-2430M CPU @ 2.400 GHz with Windows OS and 4GB of RAM.

**Table 1.** The values of the parameters in path planning examples.

| Parameter | $T_s$(sec) | $\alpha$ | $\beta$ | $\gamma$ | $\eta$ | $\zeta$ |
|---|---|---|---|---|---|---|
| value | 4 | 1.1 | 0.1 | 0.1 | 0.5 | 0.6 |

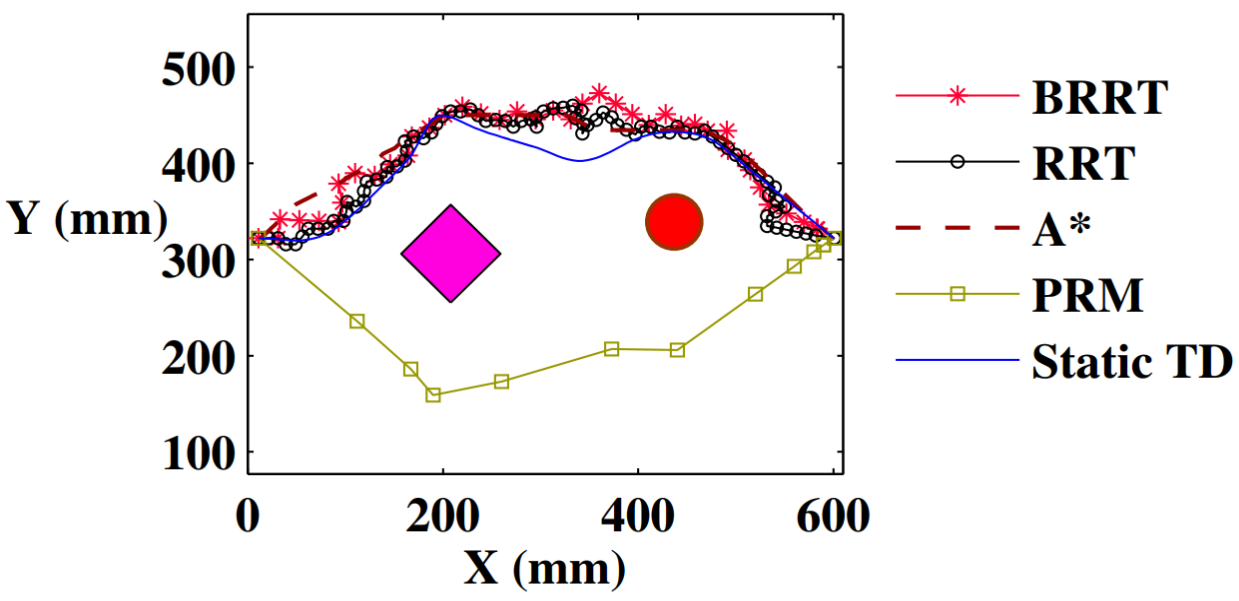

**Fig. 17.** A static path planning scenario solved by Static TD (length = 698 mm, average computational time = 0.058 s), A* (679 mm, 590 s), PRM (697 mm, 3.3 s), RRT (863 mm, 3.7 s) and BRRT (783 mm, 1.5 s). Static TD is the fastest method, and after A* and PRM, the path obtained from the static TD method is the shortest path. Subsequently, the Dynamic TD method will be used for this scenario which will result in a path with shorter length due to using TD functions in its formulation (Fig. 23).

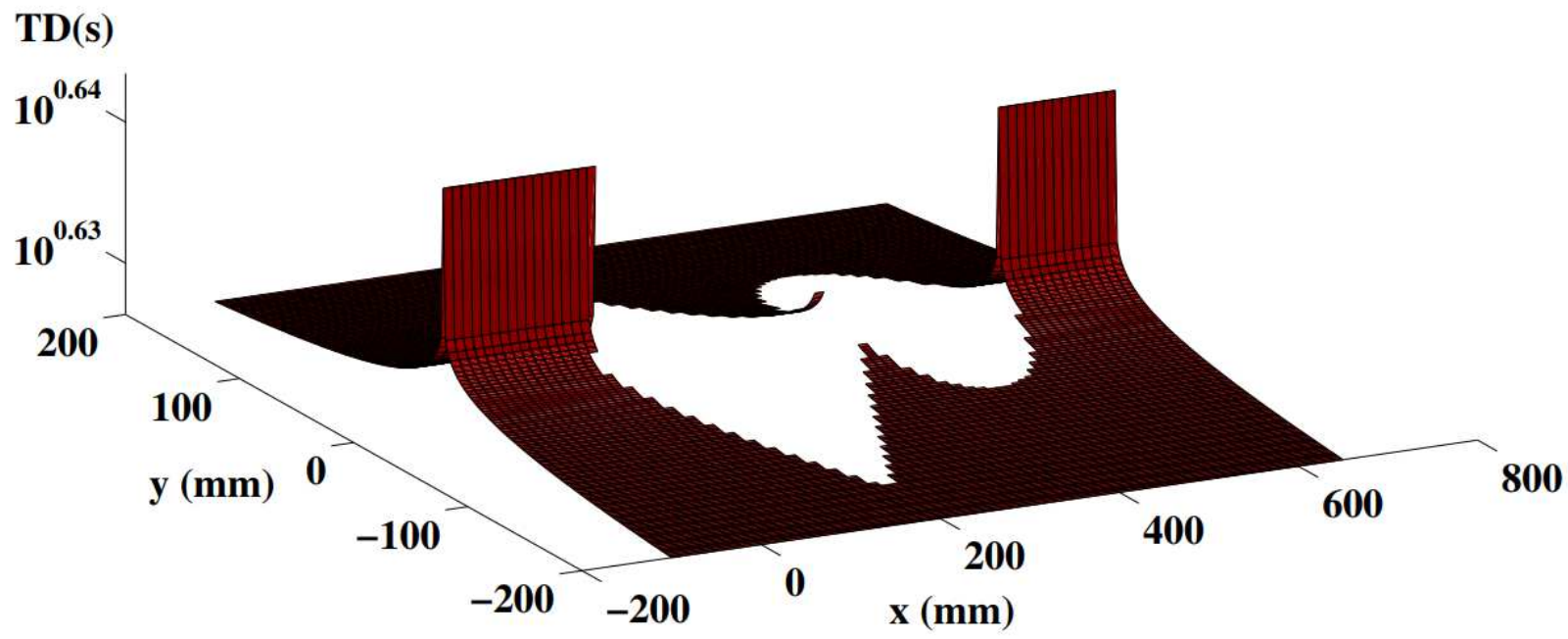

**Fig. 18.** Diagram of $TD_{conf}$ for the scenario of Fig. 17 at the first moment. Note that to generate this figure, $TD_{conf}$ is calculate over the entire configuration space. Although it seems that the obstacles are intersected after extending their dimensions because of shrinking the robot, they do not intersect in practice. This has occurred because of the rough grid chosen for this figure.

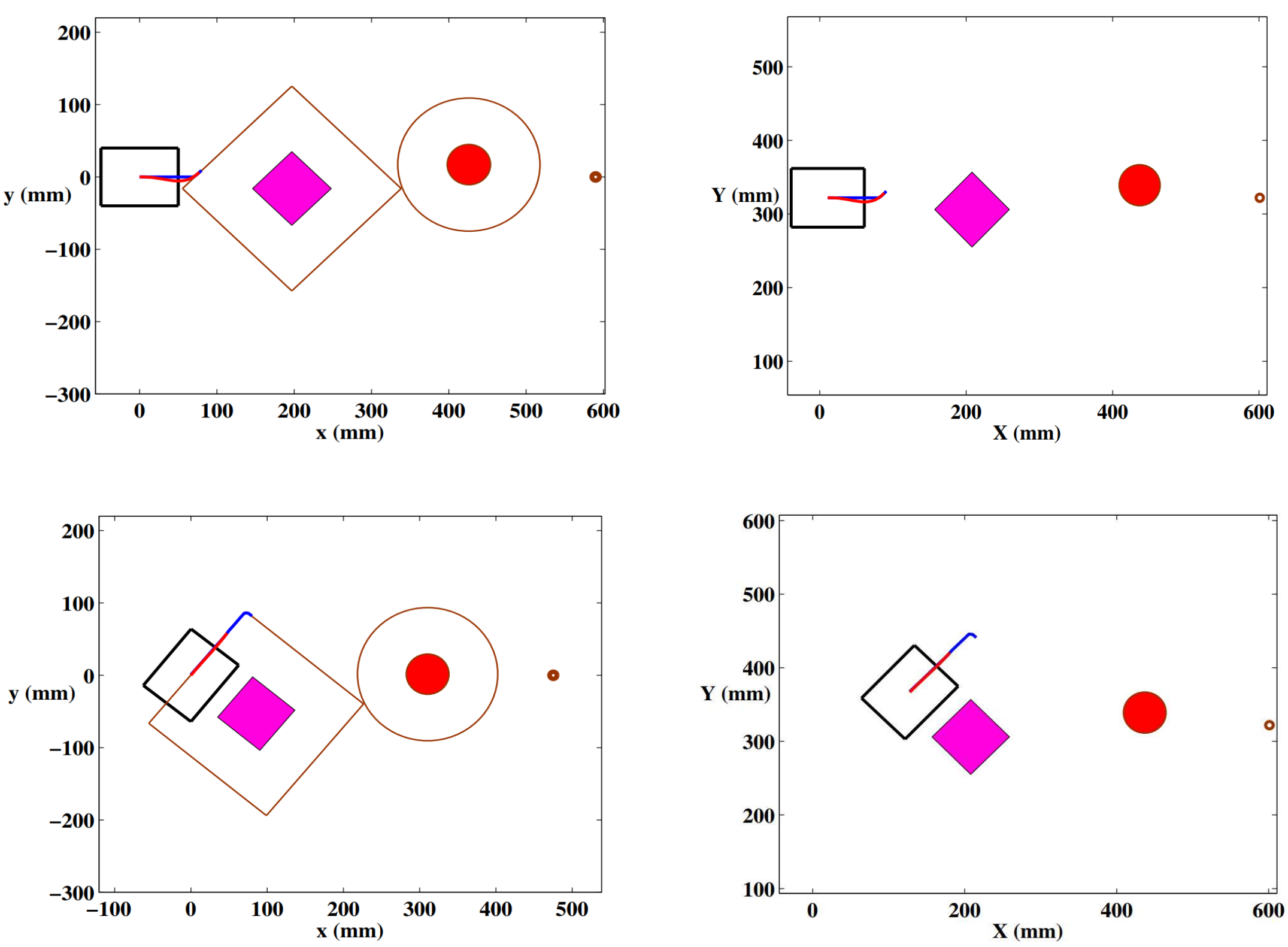

**Fig. 19.** Diagram of the planned path (blue) and trajectory (red) at some sequences of the vehicle's motion for the scenario of Fig. 17 using the Static TD method. Left: configuration space. Right: work space.

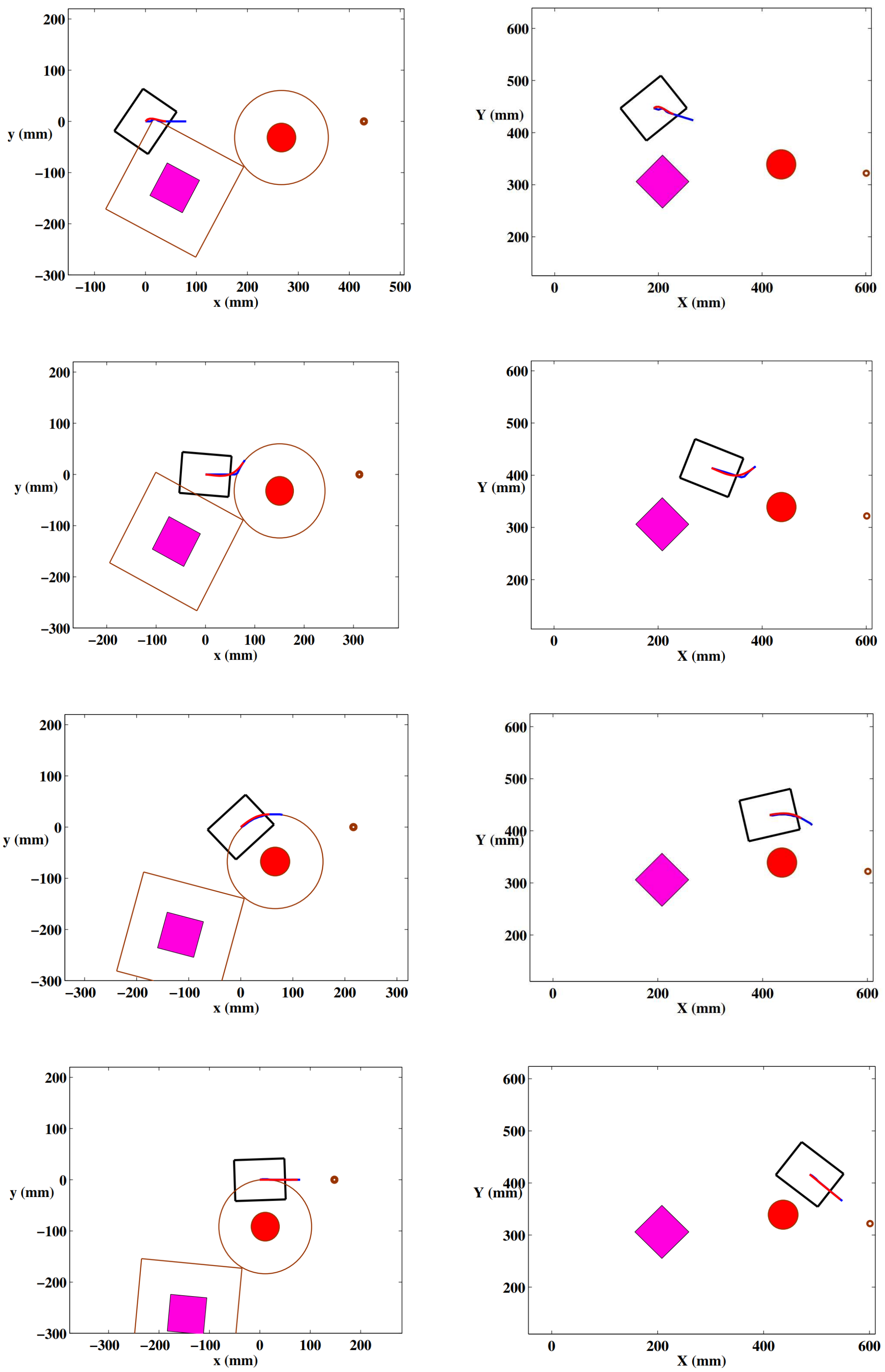

**Fig. 19.** Continued.

The Static TD method is also evaluated in cluttered environments. Figure 20 illustrates a cluttered environment and the vehicle's tracked trajectory from the start point to the goal. For this example, the average computational time is 0.064 seconds. The size of the robot and the parameters are as same as the previous example.

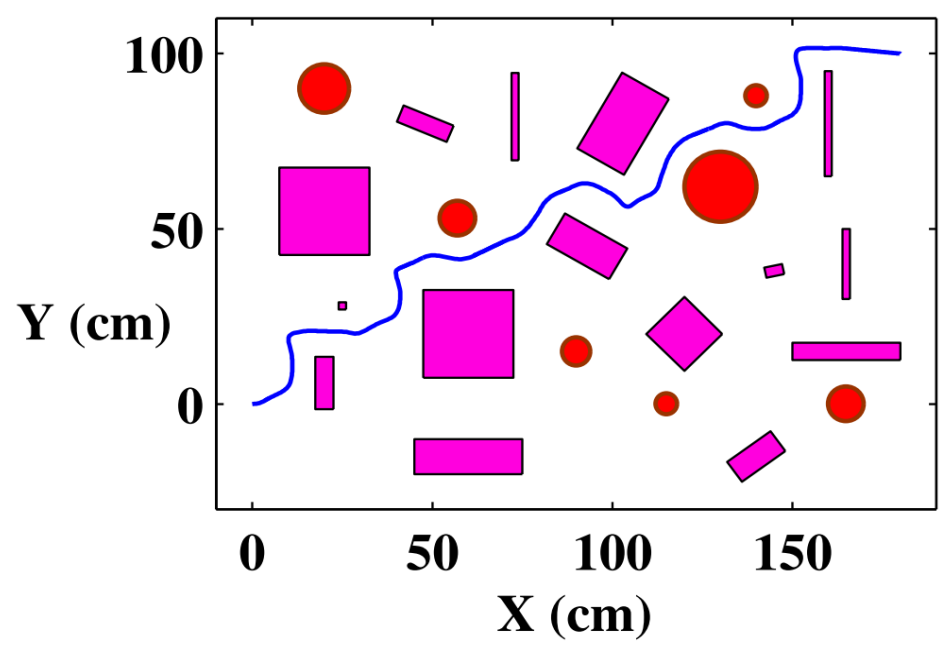

**Fig. 20.** Evaluating the Static TD method in a cluttered environment.

### 3.7 The Dynamic TD method

Since there are mobile obstacles in dynamically changing environments, it is required to predict the future state of such environments. This fact makes path planning in dynamic environments more challenging.

In dynamically changing environments, after the obstacle borders are extended to reach configuration obstacles, their future relative location and geometry with respect to the vehicle must be determined. By the future relative geometry we mean that since the TD of the vehicle with respect to different points of an obstacle is different, a relative geometry is obtained for obstacles' future location with respect to the vehicle.

To this end, first, the future relative location of vertices must be obtained. Then, the future relative location of the center of every line segment, $O_i$ , is obtained from Eq. (18) and Eq. (19), and the new length and orientation of every line segment are obtained from Eq. (20) and Eq. (33), respectively.

The TD to calculate the future relative location of every vertex point is the TD of the vertex point with respect to the axis $y_\upsilon$ . Equation (47) expresses the TD of the point $q_{R_i}$ with respect to the axis $y_\upsilon$ . In this equation, $x_{\upsilon_{q_{R_i}}}$ is the $x_\upsilon$ component of the point $q_{R_i}$ , and $\upsilon_{i_{x_\upsilon}/\upsilon}$ is the velocity of the line segment $O_i$ with respect to the vehicle in $x_\upsilon$ direction. Components of the future relative location of this point are obtained from Eq. (48). In this equation, $\upsilon_{i_x/GC}$ and $\upsilon_{i_y/GC}$ are the components of the velocity vector of the line segment $O_i$ with respect to GC (i.e. the absolute velocity vector) in the $x$ and $y$ (not X and Y ) directions, respectively. By changing the indices $q_{R_i}$ to $q_{L_i}$ in Eq. (47) and Eq. (48), similar expressions for calculating the future relative location of the vertex point $q_{L_i}$ are obtained. Note that since the future relative geometry of circular objects is elliptical, and no equation for elliptical obstacles is obtained yet, in dynamic environments, moving obstacles can only be modeled by polygons.

Finally, the path is obtained from Eq. (49) which is a combination of Eq. (43) and Eq. (38). We call this formulation, the "Dynamic TD" method. It is worth emphasizing that $Z_\infty$ and TD functions are only applied in the future relative location and geometry of configuration obstacles. The pseudo code of this algorithm is presented in Appendix 2.

$$TD_{q_{R_i}/y_\upsilon} = 2\left(1-\text{sign}\left(\frac{x_{\upsilon_{q_{R_i}}}}{\upsilon_{i_{x_\upsilon}/\upsilon}}\right)\right)^{-1}\left|\frac{x_{\upsilon_{q_{R_i}}}}{\upsilon_{i_{x_\upsilon}/\upsilon}}\right| \tag{47}$$

$$\begin{cases} x^F_{q_{R_i}} = x_{q_{R_i}} + \upsilon_{i_x/GC} \times TD_{q_{R_i}/y_\upsilon} \\ y^F_{q_{R_i}} = y_{q_{R_i}} + \upsilon_{i_y/GC} \times TD_{q_{R_i}/y_\upsilon} \end{cases} \tag{48}$$

$$\begin{aligned} y_P(x) = TD^{-1}&\left(\max_{y_{min}}^{y_{max}}\left(\min_{k=1}^{n}\left(TD_{O_k/xy}(x,y); RF(x,y); Z_{\infty_{obs_1}}(x,y), Z_{\infty_{obs_2}}(x,y),\ldots\right)\right)\right. \\ &\left.,\max_{y_{min}}^{y_{max}}\left(\min_{k=1}^{n}\left(TD_{O_k/xy}(x,y); RF(x,y); Z_{\infty_{obs_1}}(x,y), Z_{\infty_{obs_2}}(x,y),\ldots\right)\right) \geq T_s\right) \end{aligned} \tag{49}$$

#### 3.7.1 Simulation results

To evaluate the Dynamic TD method, we first solve the path planning problem of Fig. 17. Diagram of $TD_{conf}$ (over the entire configuration space) and its contour plot at the first moment of this scenario are shown in Fig. 21. Figure 22 illustrates the work space, the configuration space and contour plot of $TD_{conf}$ for this example. Variations of the path and the way TD functions affect the path during the vehicle's motion are observed in this figure.

The parameters used for solving this example using the Dynamic TD method have the same values indicated in Table 1, the path smoothing module uses the SCR-Normalize method [40] and the vehicle's longitudinal velocity is considered to be $15mm/s$ . The whole vehicle's tracked trajectory is illustrated in Fig. 23. The length of the trajectory is $678mm$ , which is $20mm$ less than the path obtained in Fig. 17 using the Static TD method (Eq. (46)) and $1mm$ shorter than the path obtained from the A* method. This fact reveals the role of TD functions in length optimization.

As a scenario in a dynamically changing environment, consider Fig. 24 in which the robot must reach the goal while avoiding collision with a mobile obstacle. In this figure, the future relative location and geometry of the mobile obstacle is also illustrated. The vehicle is a $30cm \times 20cm$ mobile robot. The path is obtained from Eq. (49) in which RF is expressed by Eq. (45). The parameters have the same values presented in Table 1, and the vehicle's longitudinal velocity is considered to be $20cm/s$ . The geometric center of the vehicle is considered as the vehicle's reference point, and the configuration obstacle is obtained by extending the lengths of the edges of the real size obstacle by the vehicle's diameter. The path smoothing module generates a 5-degree polynomial curve from the vehicle's reference point to the look-ahead point using the SCR-Normalize method [40]. When 10% of the smooth trajectory is tracked by the vehicle, the path is replanned. Similar to the previous example, considering an ideal tracking, the vehicle's motion is simulated. Figure 25 illustrates the diagram of $TD_{conf}$ (over the entire configuration space) and its contour plot at the first moment of this scenario. Figure 26 demonstrates the work space, the configuration space and contour plot of $TD_{conf}$ during this simulation. Figure 27 illustrates the whole trajectory from the start point to the goal. The average computational time for this example is 0.051 seconds.

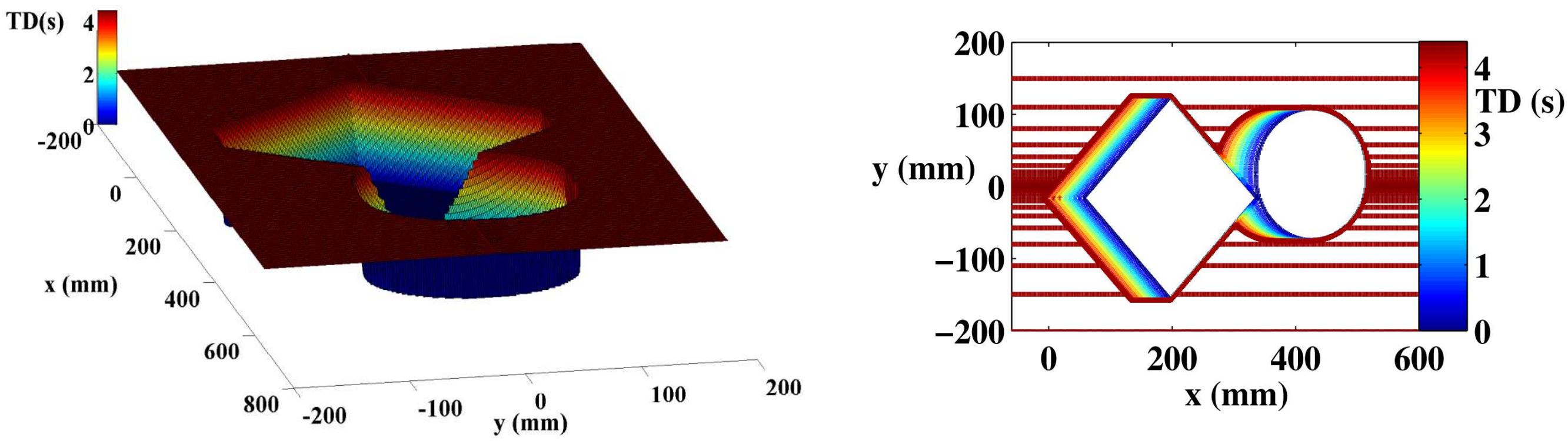

**Fig. 21.** Diagram of $TD_{conf}$ for the scenario of Fig. 17 at the first moment (left) and its contour plot (right). Note that to generate this figure, $TD_{conf}$ is calculated over the entire configuration space.

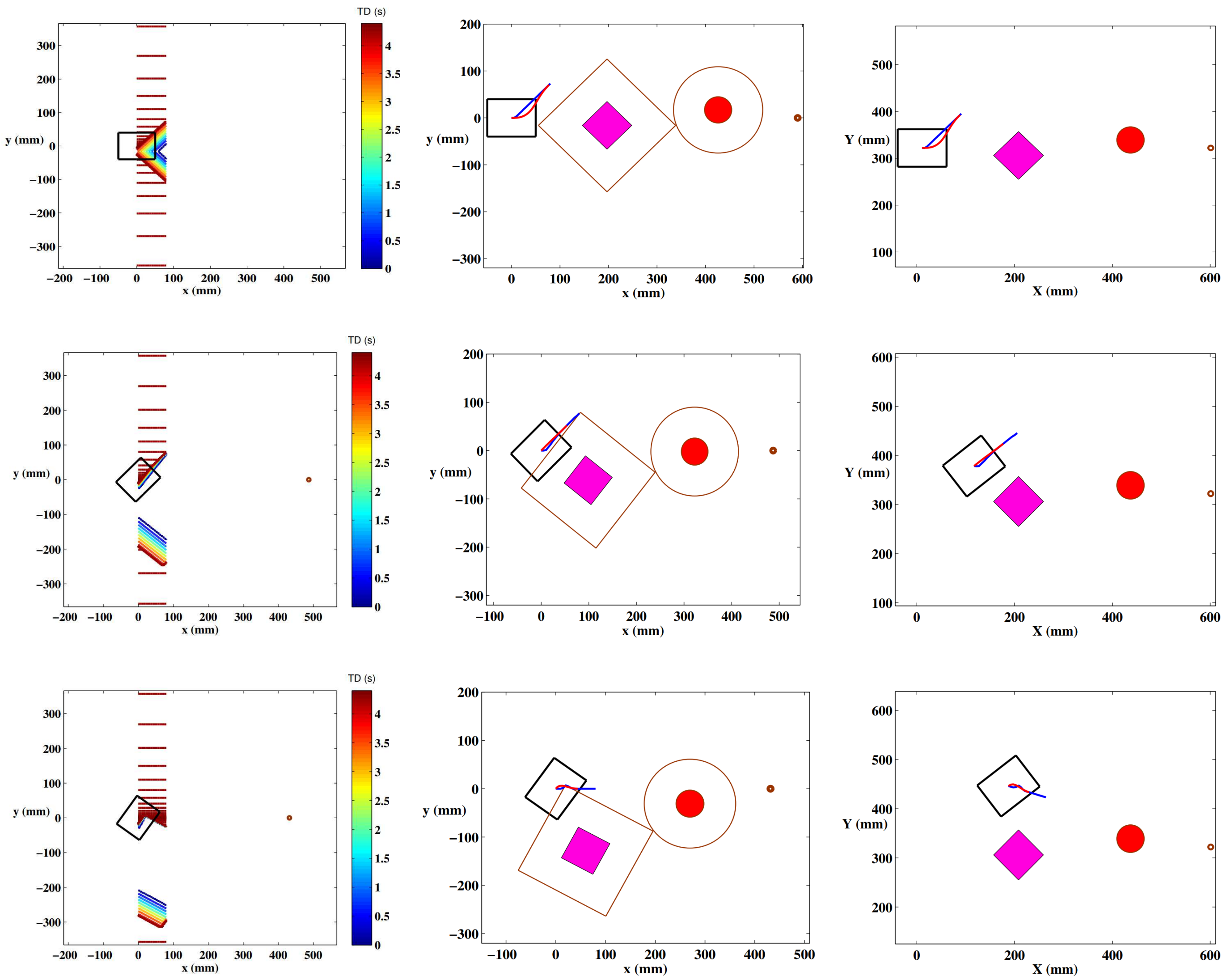

**Fig. 22.** Diagram of the planned path (blue) and trajectory (red) at some sequences of the vehicle's motion for the scenario of Fig. 17 and contour plot of $TD_{conf}$ at these sequences. The scenario is solved using the Dynamic TD method. Left: contour plot of $TD_{conf}$. Middle: configuration space. Right: work space

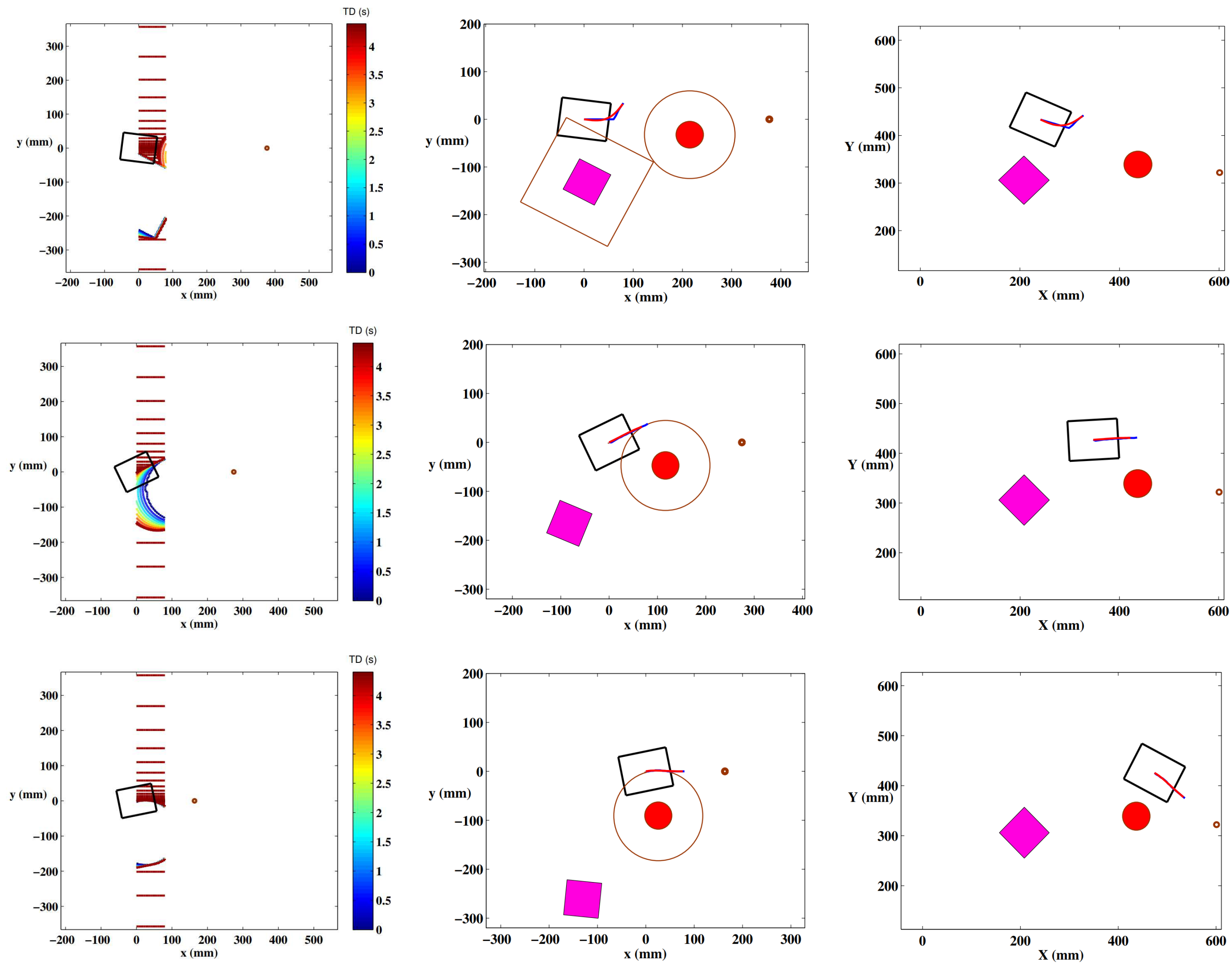

Fig. 22. Continued.

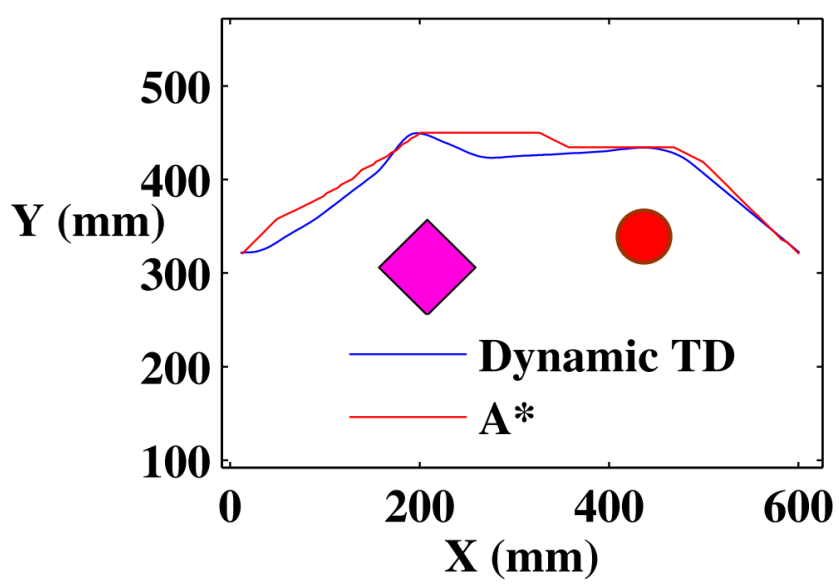

Fig. 23. Diagram of the whole vehicle's tracked trajectory for the scenario of Fig. 17 using the Dynamic TD method (path length = 678 mm, average computational time = 0.071 s). The path is shorter than the path obtained from the A* method (Fig. 17) and 20 mm shorter than the path obtained from the Static TD method demonstrating the effect of using TD functions in the path planning formulation, on the optimality of the path.

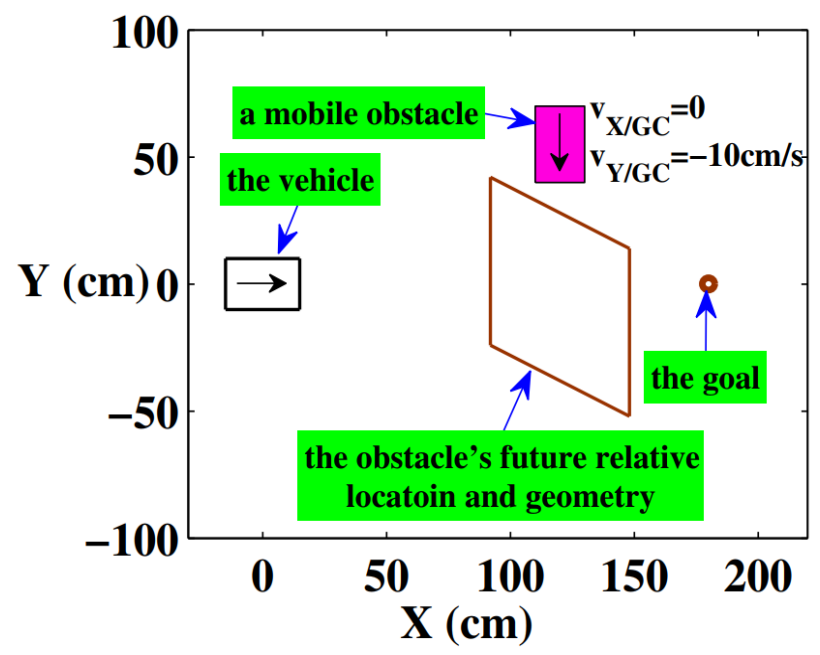

Fig. 24. A dynamic scenario for path planning: a mobile robot and its goal, and a mobile obstacle and its future relative location and geometry.

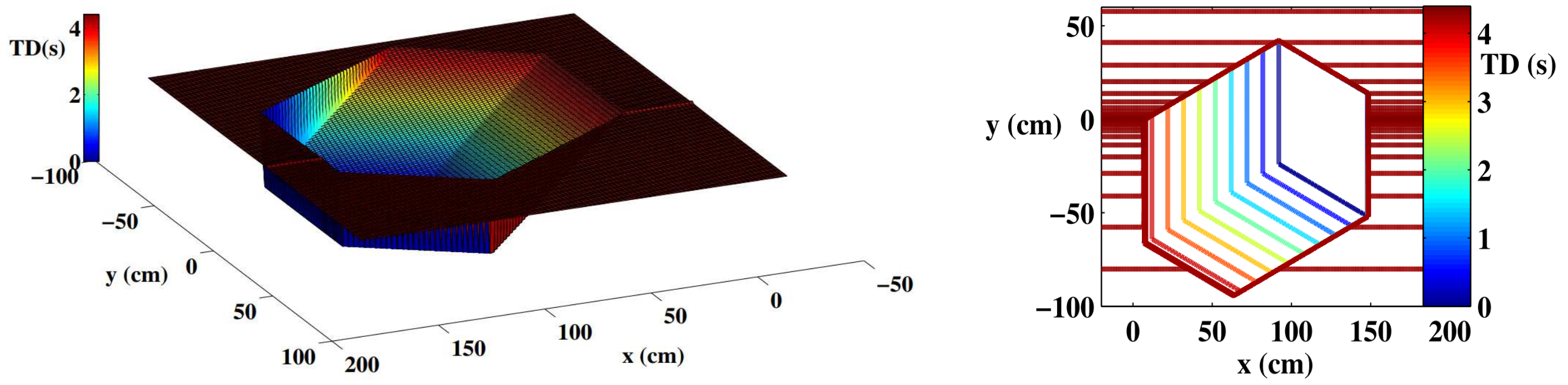

**Fig. 25.** Diagram of $TD_{conf}$ for the scenario of Fig. 24 at the first moment (left) and its contour plot (right). Note that to generate this figure, $TD_{conf}$ is calculated over the entire configuration space.

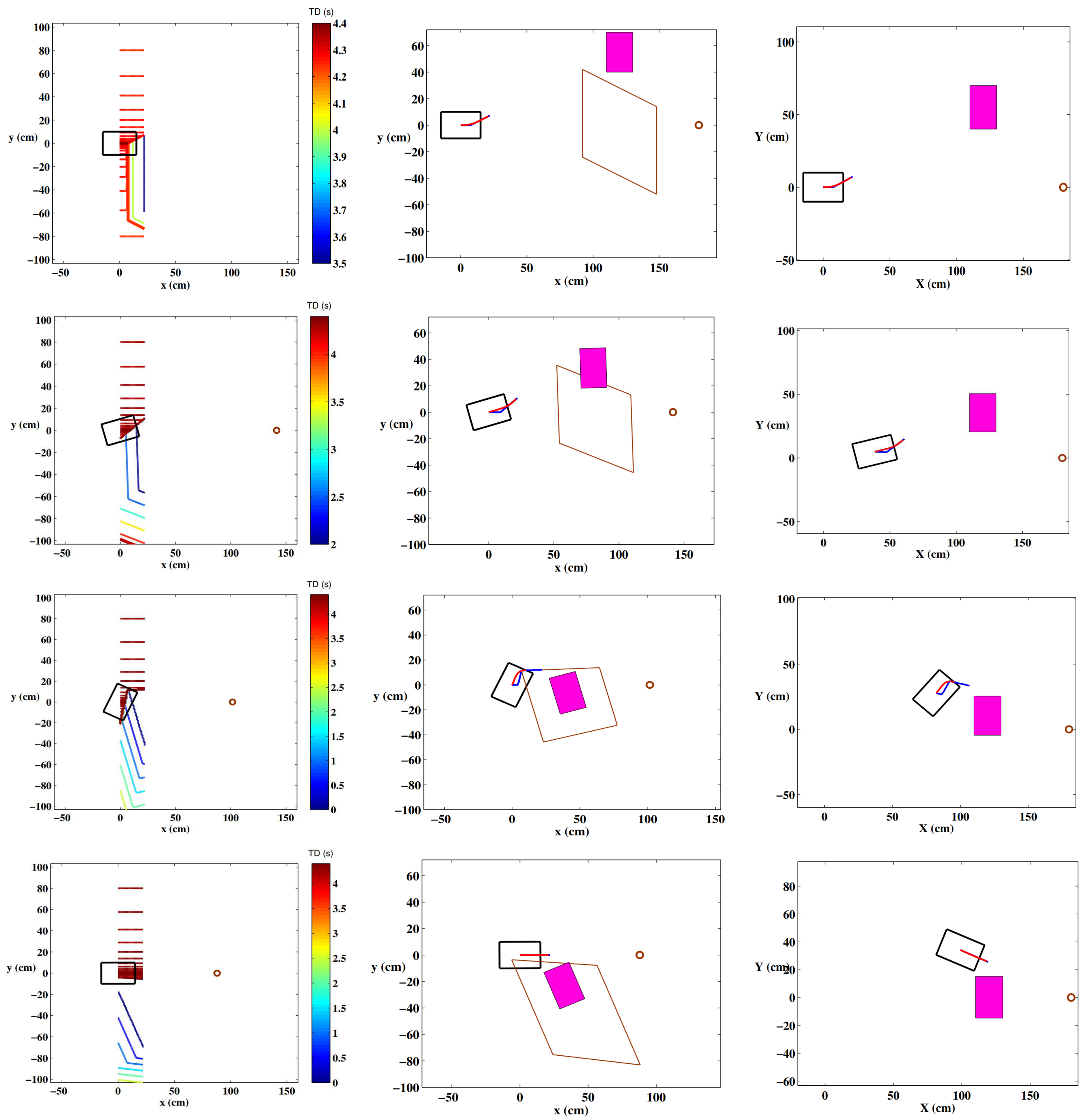

**Fig. 26.** Diagram of the planned path (blue) and trajectory (red) at some sequences of the vehicle's motion for the scenario of Fig. 24 and contour plot of $TD_{conf}$ at these sequences. The scenario is solved using the Dynamic TD method. Left: contour plot of $TD_{conf}$. Middle: configuration space. Right: work space.

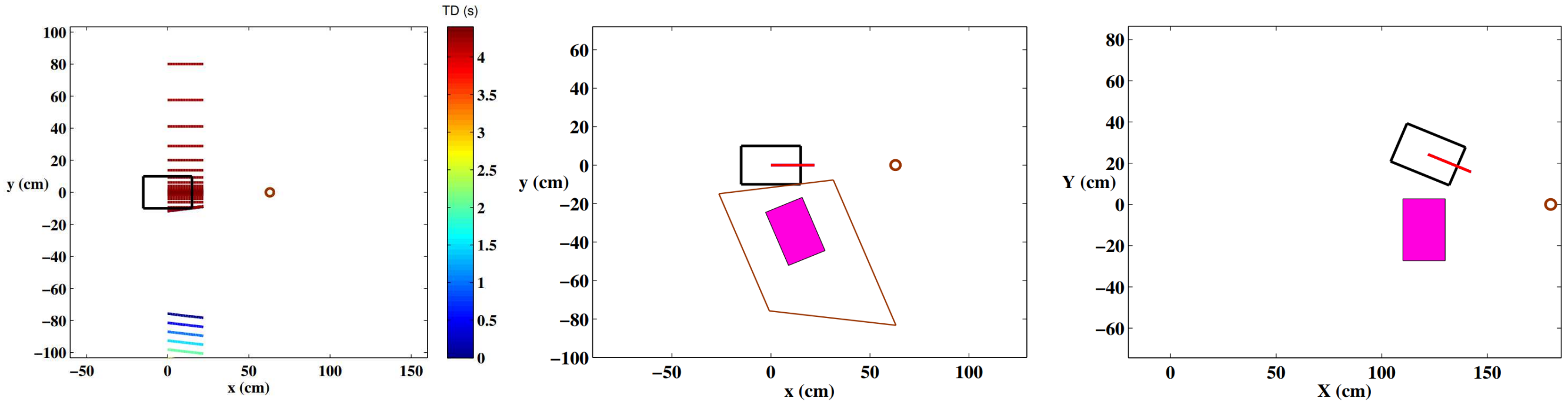

Fig. 26. Continued.

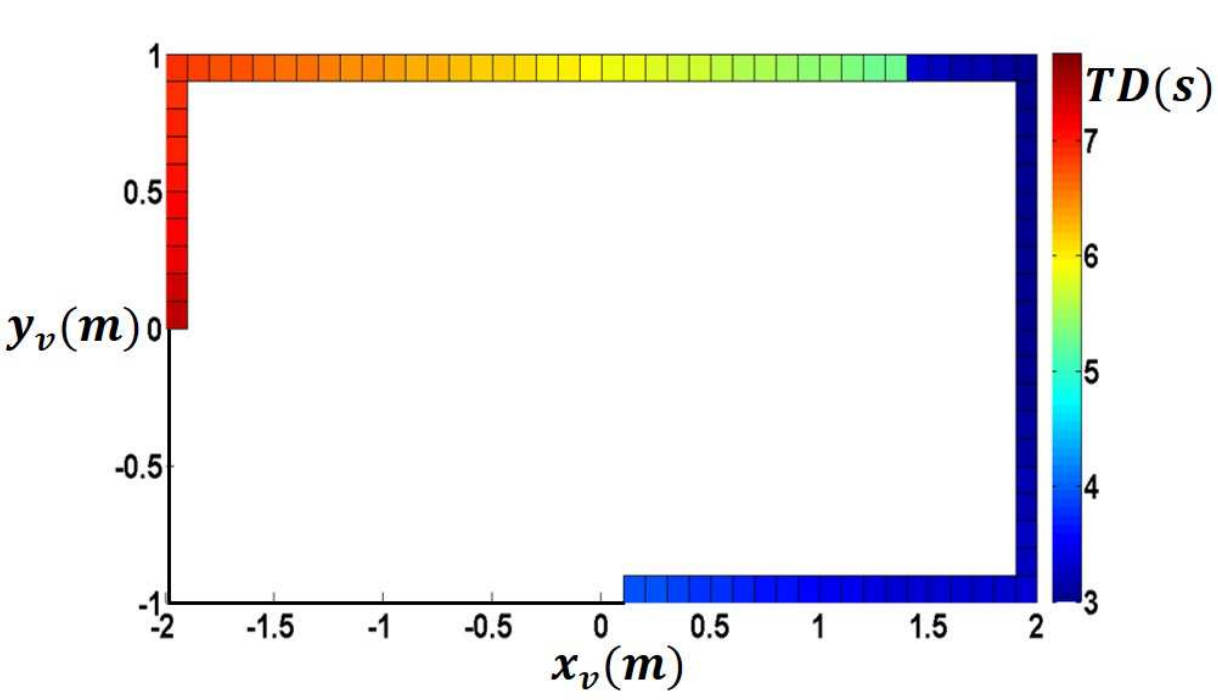

Fig. 27. Diagram of the whole vehicle's tracked trajectory for the scenario of Fig. 24 using the Dynamic TD method. The average computational time is 0.051 seconds.

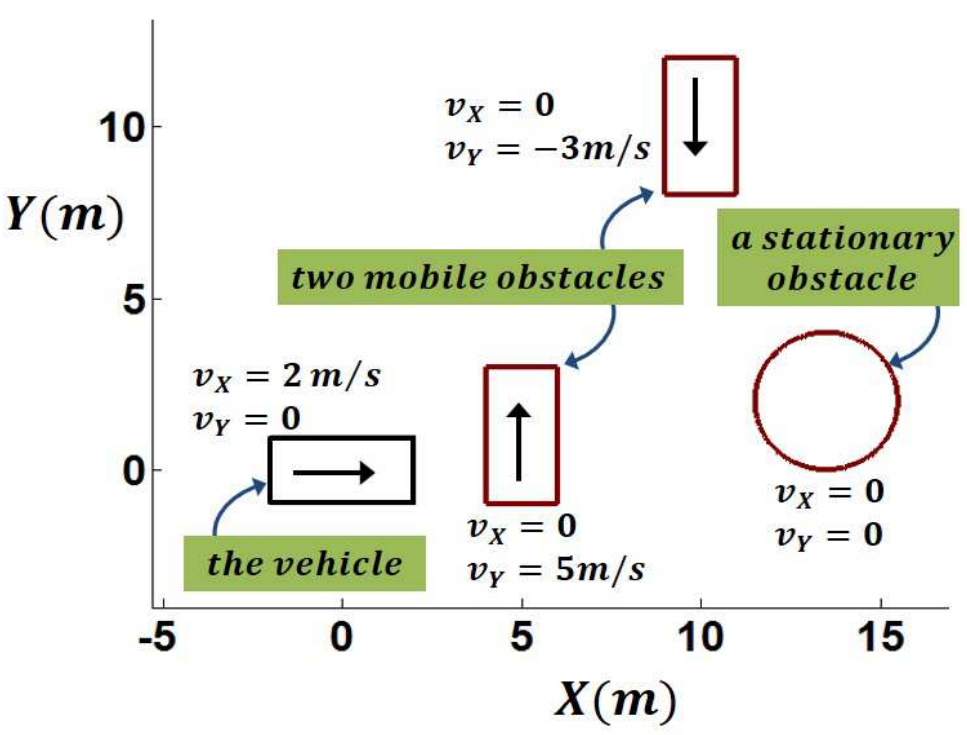

Fig. 28. A scenario for collision prediction.

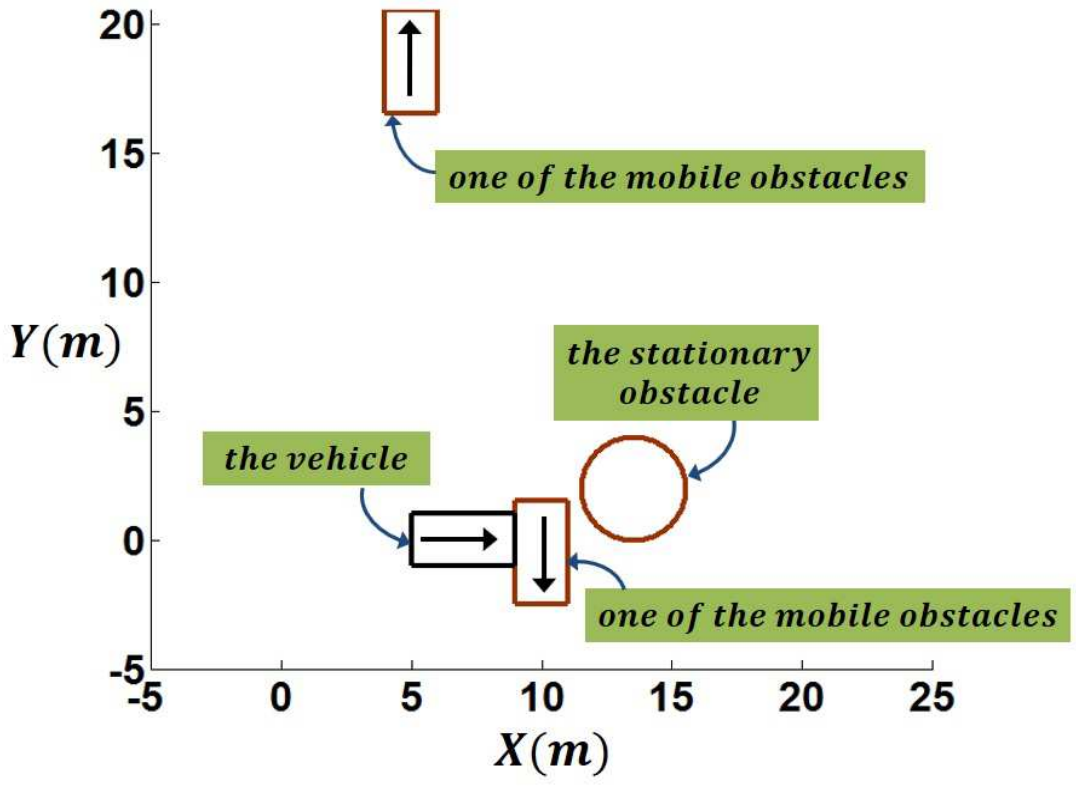

Fig. 29. Diagram of the TD of the points on the vehicle's border with respect to the surrounding obstacles.

Fig. 30. Location of the obstacles and the vehicle after 3.5 seconds.

## 4. Collision Prediction

The semi-automated vehicles in which the driver assistance system is automatically activated when the imminence of a collision is predicted, require a collision prediction module. In addition, many motion planning methods have a collision checking/detection step. TD functions can be used to construct a collision prediction framework for activating driver assistance systems or function as a collision checking/detection module in some motion planning frameworks. Note that as it is evident from the previous sections, the TD path planning framework itself does not require such a module.

In this section, we present a collision prediction framework based on the TD concept. According to the definition of TD, TD of the object A with respect to the object B is equal to TD of B with respect to A. Therefore, it is concluded that TD of a point $u(x_u, y_u)$ in the space $xy$ with respect to a set of geometric objects G (defined by Eq. (9)) is equal to the TD of G with respect to the space $xy$ at the point $u(x_u, y_u)$. Therefore, we have:

$$TD_{u/G} = TD_{G/xy}(x_u, y_u) = \min_{k=1}^{n}\left(TD_{O_k/xy}(x_u, y_u)\right) \tag{50}$$

To predict collision and the time to collision of a vehicle with its surrounding obstacles, the TD of the points on the vehicle's border with respect to the set of all surrounding obstacles must be obtained. A finite number of points for this purpose suffices; for example, four corner points of the vehicle. To carry this out, the location and velocity of all obstacles must be expressed in a coordinate reference on the vehicle (i.e. its velocity is equal to the vehicle's velocity). Then, Eq. (50) is used to determine the TD of every point u on the border of the vehicle with respect to the set of the surrounding obstacles.

The minimum value of the TD of the points on the vehicle's border with respect to the surrounding obstacles is the time remaining for the vehicle to be involved in a collision (time to collision). If the value of this minimum is infinity, no collision will occur between the vehicle and its surrounding obstacles. Note that it is also possible to determine the obstacle with which the vehicle will collide by obtaining the mentioned minimum value for every obstacle separately.

### 4.1 Simulation results

Figure 28 illustrates a vehicle and its surrounding obstacles. Diagram of the TD of the points on the vehicle's border with respect to the surrounding obstacles is shown in Fig. 29. The time to collision is obtained 3.5s. Therefore, if the velocities of all obstacles and the vehicle do not change, the vehicle will collide with at least one of the obstacles in 3.5 seconds. To validate this prediction, all obstacles are indicated 3.5 seconds later in Fig. 30. Note that, for this example, VC is used for calculating the TDs and GC is used for illustrating the work space.

## 5. Conclusions and Future Work

In this paper, the TD framework for path planning in two dimensional dynamic environments was introduced. As a local planner, this method exploits the idea of space-time space to generate a length optimal collision free path in a dynamic environment. The most important feature of this framework is to use an explicit formula to generate the path. This feature accounts for the quickness of the TD method in comparison to the methods which require hierarchical operations such as sampling, collision checking, searching, etc. Furthermore, the path obtained by the TD method is well optimized in terms of path length. This fact was illustrated in a comparative example. In addition, this framework benefits from an inherent collision prediction in the time dimension that provides a spectrum of low-risk to high-risk regions in the configuration-time space. This characteristic leads to the ability to have a desired distance in the time dimension from obstacles, which results in a good performance in collision avoidance.

Another achievement of this research was presenting a collision prediction algorithm based on the TD concept. This algorithm can be helpful in semi-automated vehicles and other path planning frameworks.

There are abundant future works associated with the TD method. Obtaining the TD function for elliptical shape objects and implementing the future relative location and geometry for circular and elliptical objects is of importance. In addition, formulating this framework for path planning of automated ground vehicles is a useful future work. Furthermore, determining $T_s$ based on the vehicle's maneuvering capabilities and road-vehicle parameters will be a beneficial research. Finally, implementing the TD method to multi-agent systems is one of our future plans.

## Author Contributions

As supervisors, S. Azadi and R. Kazemi designed and directed the project. A. Analooee developed the theories, performed the simulations and wrote the manuscript. All authors discussed the results, reviewed, and approved the final version of the manuscript.

## Acknowledgments

Not Applicable.

## Conflict of Interest

The authors declared no potential conflicts of interest concerning the research, authorship, and publication of this article.

## Funding

The authors received no financial support for the research, authorship, and publication of this article.

## Data Availability Statements

The datasets generated and/or analyzed during the current study are available from the authors on reasonable request.

## Appendix 1: Pseudo Code of Static TD Algorithm

In this appendix, the pseudo code of the Static TD algorithm is presented. The input to this algorithm is the obstacles matrix M which is considered to contain the required features of the obstacles. It is also considered that the obstacles are enlarged to turn into configuration obstacles.

**Input:** Obstacles Matrix M
**Output:** Desired Trajectory
$[x,y]=\text{meshgrid}[0<x<\zeta D,\ y_{min}<y<y_{max}]$
$TD_{conf}=+\infty$
Σ=number of obstacles
For $\sigma$ = 1, 2, …, Σ:
    $Z_{\infty_\sigma}=0$
    If obstacle σ is a polygon:
        m= number of edges of the polygon
        For i= 1, 2, …, m:

$$x_{O_i}^0=\frac{x_{q_{Ri}}+x_{q_{Li}}}{2}$$

$$y_{O_i}^0 = \frac{y_{q_{Ri}} + y_{q_{Li}}}{2}$$

$$\theta_i'' = \tan^{-1}\left(\frac{y_{q_{Ri}} - y_{q_{Li}}}{x_{q_{Ri}} - x_{q_{Li}}}\right) + \frac{\pi}{2}\mathrm{sign}\left(\mathrm{sign}\left(x_{q_{Ri}} - x_{q_{Li}}\right) + 0.5\right)$$

$$\begin{bmatrix} x_i'' \\ y_i'' \end{bmatrix} = \begin{bmatrix} \cos\theta_i'' & \sin\theta_i'' \\ -\sin\theta_i'' & \cos\theta_i'' \end{bmatrix}\begin{bmatrix} x \\ y \end{bmatrix}$$

$$\begin{bmatrix} x_{iO_i}''^0 \\ y_{iO_i}''^0 \end{bmatrix} = \begin{bmatrix} \cos\theta_i'' & \sin\theta_i'' \\ -\sin\theta_i'' & \cos\theta_i'' \end{bmatrix}\begin{bmatrix} x_{O_i}^0 \\ y_{O_i}^0 \end{bmatrix}$$

$$Z_{\infty_{O_i}}(x,y) = \left(\mathrm{sign}\left(\mathrm{sign}\left(x_{iO_i}''^0 - x_i''\right) + 1\right)\right)^{-1} - 1$$

$$Z_{\infty_\sigma}(x,y) = \max\left(Z_{\infty_\sigma}(x,y), Z_{\infty_{O_i}}(x,y)\right)$$

$$TD_{conf} = \min\left(TD_{conf}, Z_{\infty_\sigma}(x,y)\right)$$

If obstacle $\sigma$ is a circle:

$$Z_{\infty_\sigma} = \left(\mathrm{sign}\left(\mathrm{sign}\left(R^2 - (x - x_C)^2 - (y - y_C)^2\right) + 1\right)\right)^{-1} - 1$$

$$TD_{conf} = \min\left(TD_{conf}, Z_{\infty_\sigma}(x,y)\right)$$

$$RF(x,y) = \left(\mathrm{sign}\left(\mathrm{sign}\left(y_P(L)\right)\left\lfloor\frac{|y_P(L)|}{\eta D}\right\rfloor\right)\mathrm{sign}(y) + \left|\mathrm{sign}\left(\left\lfloor\frac{|y_P(L)|}{\eta D}\right\rfloor\right) - 1\right|\right)\left(\alpha T_s - \beta\,|\,y\,|^{\gamma}\right)$$

$$t_P(x) = \max_{y_{min}}^{y_{max}}\left(\min\left(TD_{conf}(x,y); RF(x,y)\right)\right)$$

if $t_P(x) > 0$:

$$y_P(x) = TD^{-1}\left(t_P(x)\right)$$

else:

Break

Smooth the path by generating a Trajectory from $(0,0)$ to $(L, y_P(L))$

**Return** Trajectory

## Appendix 2: Pseudo Code of Dynamic TD Algorithm

In this appendix, the pseudo code of the Dynamic TD algorithm is presented. The input to this algorithm is the obstacles matrix M which is considered to contain the required features of the obstacles. It is also considered that the obstacles are enlarged to turn into configuration obstacles.

**Input:** Obstacles Matrix M

**Output:** Desired Trajectory

[x,y]=meshgrid[0<x< $\zeta D$ , $y_{min}$<y<$y_{max}$]

$TD_{conf} = +\infty$

Σ=number of obstacles

For $\sigma$ = 1, 2, …, Σ:

$Z_{\infty_\sigma} = 0$

If obstacle $\sigma$ is a polygon:

m= number of edges of the polygon

For i= 1, 2, …, m:

$$TD_{q_{R_i}/y_v} = 2\left(1 - \mathrm{sign}\left(\frac{x_{v_{qR_i}}}{v_{i_{x_v}/v}}\right)\right)^{-1}\left|\frac{x_{v_{qR_i}}}{v_{i_{x_v}/v}}\right|$$

$$TD_{q_{L_i}/y_v} = 2\left(1 - \mathrm{sign}\left(\frac{x_{v_{qL_i}}}{v_{i_{x_v}/v}}\right)\right)^{-1}\left|\frac{x_{v_{qL_i}}}{v_{i_{x_v}/v}}\right|$$

$$\begin{cases} x_{q_{R_i}}^F = x_{q_{R_i}} + v_{i_x/GC} \times TD_{q_{R_i}/y_v} \\ y_{q_{R_i}}^F = y_{q_{R_i}} + v_{i_y/GC} \times TD_{q_{R_i}/y_v} \end{cases}$$

$$\begin{cases} x_{q_{L_i}}^F = x_{q_{L_i}} + v_{i_x/GC} \times TD_{q_{L_i}/y_v} \\ y_{q_{L_i}}^F = y_{q_{L_i}} + v_{i_y/GC} \times TD_{q_{L_i}/y_v} \end{cases}$$

$$x_{O_i}^0 = \frac{x_{q_{Ri}}^F + x_{q_{Li}}^F}{2}$$

$$y_{O_i}^0 = \frac{y_{q_{Ri}}^F + y_{q_{Li}}^F}{2}$$

$$\theta_i'' = \tan^{-1}\left(\frac{y^F_{q_{Ri}} - y^F_{q_{Li}}}{x^F_{q_{Ri}} - x^F_{q_{Li}}}\right) + \frac{\pi}{2}\mathrm{sign}\left(\mathrm{sign}\left(x^F_{q_{Ri}} - x^F_{q_{Li}}\right) + 0.5\right)$$

$$\begin{bmatrix} x_i'' \\ y_i'' \end{bmatrix} = \begin{bmatrix} \cos\theta_i'' & \sin\theta_i'' \\ -\sin\theta_i'' & \cos\theta_i'' \end{bmatrix} \begin{bmatrix} x \\ y \end{bmatrix}$$

$$\begin{bmatrix} x''^0_{iO_i} \\ y''^0_{iO_i} \end{bmatrix} = \begin{bmatrix} \cos\theta_i'' & \sin\theta_i'' \\ -\sin\theta_i'' & \cos\theta_i'' \end{bmatrix} \begin{bmatrix} x^0_{O_i} \\ y^0_{O_i} \end{bmatrix}$$

$$\theta_i' = \tan^{-1}\left(\frac{\upsilon_{i_y/xy}}{\upsilon_{i_x/xy}}\right) - \frac{\pi}{2}\left(\mathrm{sign}\left(\upsilon_{i_x/xy}\right) - 1\right)\mathrm{sign}\left(\upsilon_{i_x/xy}\right)$$

$$l_i = \sqrt{\left(x^F_{q_{Ri}} - x^F_{q_{Li}}\right)^2 + \left(y^F_{q_{Ri}} - y^F_{q_{Li}}\right)^2}$$

$$y''_{iO_i}\left(x_i''\right) = y''^0_{iO_i} + \left(x_i'' - x''^0_{iO_i}\right)\tan(\theta_i' - \theta_i'')$$

$$Q_i = \mathrm{sign}\left(\mathrm{sign}\left(\frac{l_i}{2} - \left|y_i'' - y''_{iO_i}(x_i'')\right|\right) + 1\right)$$

$$TD_{O_i/xy}(x,y) = 2Q_i^{-1}\left(\mathrm{sign}\left(\frac{x_i'' - x''^0_{iO_i}}{\upsilon_{x_i'O_i/xy}}\right) + 1\right)^{-1}\left|\frac{x_i'' - x''^0_{iO_i}}{\upsilon_{x_i'O_i/xy}}\right|$$

$$TD_{conf} = \min\left(TD_{conf}, TD_{O_i/xy}(x,y)\right)$$

$$Z_{\infty_{O_i}}(x,y) = \left(\mathrm{sign}\left(\mathrm{sign}\left(x''^0_{iO_i} - x_i''\right) + 1\right)\right)^{-1} - 1$$

$$Z_{\infty_\sigma}(x,y) = \max\left(Z_{\infty_\sigma}(x,y), Z_{\infty_{O_i}}(x,y)\right)$$

$$TD_{conf} = \min\left(TD_{conf}, Z_{\infty_\sigma}(x,y)\right)$$

If obstacle $\sigma$ is a circle (currently, only stationary obstacles can be modeled as circles):

$$\theta_\sigma' = \tan^{-1}\left(\frac{\upsilon_{\sigma_y/xy}}{\upsilon_{\sigma_x/xy}}\right) - \frac{\pi}{2}\left(\mathrm{sign}\left(\upsilon_{\sigma_x/xy}\right) - 1\right)\mathrm{sign}\left(\upsilon_{\sigma_x/xy}\right)$$

$$\begin{bmatrix} x_\sigma' \\ y_\sigma' \end{bmatrix} = \begin{bmatrix} \cos\theta_\sigma' & \sin\theta_\sigma' \\ -\sin\theta_\sigma' & \cos\theta_\sigma' \end{bmatrix} \begin{bmatrix} x \\ y \end{bmatrix}$$

$$\begin{bmatrix} x'^0_{\sigma O_\sigma} \\ y'^0_{\sigma O_\sigma} \end{bmatrix} = \begin{bmatrix} \cos\theta_\sigma' & \sin\theta_\sigma' \\ -\sin\theta_\sigma' & \cos\theta_\sigma' \end{bmatrix} \begin{bmatrix} x^0_{O_\sigma} \\ y^0_{O_\sigma} \end{bmatrix}$$

$$\upsilon_{O_\sigma/xy} = \sqrt{\upsilon^2_{\sigma_x/xy} + \upsilon^2_{\sigma_y/xy}}$$

$$Q_\sigma = \mathrm{sign}\left(\mathrm{sign}\left(R_\sigma - \left|y_\sigma' - y'^0_{\sigma O_\sigma}\right|\right) + 1\right)$$

$$TD_{O_\sigma/xy}(x,y) = 2Q_\sigma^{-1}\left(\mathrm{sign}\left(\frac{x_\sigma' - x'^0_{\sigma O_\sigma} - Q_\sigma\sqrt{R_\sigma^2 - (y_\sigma' - y'^0_{\sigma O_\sigma})^2}}{\upsilon_{O_\sigma/xy}}\right) + 1\right)^{-1} \times \left|\frac{x_\sigma' - x'^0_{\sigma O_\sigma} - Q_\sigma\sqrt{R_\sigma^2 - (y_\sigma' - y'^0_{\sigma O_\sigma})^2}}{\upsilon_{O_\sigma/xy}}\right|$$

$$Z_{\infty_\sigma} = \left(\mathrm{sign}\left(\mathrm{sign}\left(R^2 - (x - x_C)^2 - (y - y_C)^2\right) + 1\right)\right)^{-1} - 1$$

$$TD_{conf} = \min\left(TD_{conf}, TD_{O_\sigma/xy}(x,y); Z_{\infty_\sigma}(x,y)\right)$$

$$RF(x,y) = \left(\mathrm{sign}\left(\mathrm{sign}\left(y_P(L)\right)\left\lfloor\frac{|y_P(L)|}{\eta D}\right\rfloor\right)\mathrm{sign}(y) + \left|\mathrm{sign}\left(\left\lfloor\frac{|y_P(L)|}{\eta D}\right\rfloor\right) - 1\right|\right)\left(\alpha T_s - \beta\,|\,y\,|^\gamma\right)$$

$$t_P(x) = \max_{y_{min}}^{y_{max}}\left(\min\left(TD_{conf}(x,y); RF(x,y)\right)\right)$$

if $t_P(x) \geq T_s$ :

$$y_P(x) = TD^{-1}\left(t_P(x)\right)$$

else:

Break

Smooth the path by generating a Trajectory from $(0,0)$ to $(L, y_P(L))$

**Return** Trajectory

## ORCID iD

Ali Analooee https://orcid.org/0000-0001-7049-1587
Shahram Azadi https://orcid.org/0000-0002-5817-9838
Reza Kazemi https://orcid.org/0000-0003-1441-4602

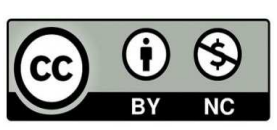